\documentclass[letterpaper,journal]{IEEEtran}
\usepackage{amsmath,amsfonts}
\usepackage{algorithmic}
\usepackage{algorithm}
\usepackage{array}
\usepackage[caption=false,font=normalsize,labelfont=sf,textfont=sf]{subfig}
\usepackage{textcomp}
\usepackage{stfloats}
\usepackage{url}
\usepackage{verbatim}
\usepackage{graphicx}
\usepackage{cite}
\usepackage{booktabs}
\usepackage{multirow}
\usepackage{makecell}

\begin{document}

\title{Learning Agile Navigation in Crowded Environments \\ for Quadruped Robots}

\author{Shuyu~Wu, Zeyu~Liu, Tianbao~Zhang, Fanxing~Li, Fangyu~Sun,
Mingkang~Xiong, Wei~Xi, Wenxian~Yu, Danping~Zou\textsuperscript{*}
\thanks{Shuyu Wu, Zeyu Liu, Tianbao Zhang, Fanxing Li, Fangyu Sun, Wenxian Yu, and Danping Zou are with Shanghai Jiao Tong University, Shanghai 200240, China (e-mail: dpzou@sjtu.edu.cn).}
\thanks{Mingkang Xiong and Wei Xi are with State Key Laboratory of High-end Heavy-load Robots, Midea Group, Foshan 528300, China.}
\thanks{\textsuperscript{*}Corresponding author: Danping Zou.}}

\maketitle

\begin{abstract}
Navigating dynamic and crowded environments presents significant challenges for quadruped robots due to severe sensor occlusion and unpredictable human motion. Existing approaches face a trade-off: model-based methods, such as Velocity Obstacles (VO), theoretically guarantee safety but rely on accurate obstacle motion estimates that often fail in dense crowds, while end-to-end learning methods offer robustness but lack motion prediction capability of obstacles, leading to collisions or conservative behaviors. To solve this, we propose VOP-Nav, a novel navigation system that combines the geometric safety of VO with the agile adaptability of end-to-end learning. Using only local onboard observations, our system avoids explicit obstacle detection and tracking pipelines. The VOP-Net processes multi-frame LiDAR data to implicitly encode dynamic constraints and predict {a} safe velocity region derived from Velocity Obstacle theory. Importantly, the VO predictions serve a dual role: they are used as input to the navigation policy during inference and as a reward signal during training to encourage safe motion. Evaluations in Isaac Gym demonstrate that VOP-Nav {achieves} higher success rates than all baselines while balancing locomotion speed and collision avoidance. Real-world deployment on a Unitree Go2 quadruped robot further validates the system's robustness and efficiency in complex indoor and outdoor dynamic environments.

\end{abstract}


\section{Introduction}

\IEEEPARstart{R}{obots} operating in everyday environments, such as urban streets, shopping centers, factories, and public transportation systems, must navigate dynamic and crowded scenes. Ensuring safe navigation in such settings remains challenging. 
Dynamic obstacles often cause sensor occlusion, degrading perception quality, while the motion of surrounding agents, such as pedestrians and vehicles, is often difficult to predict, particularly in densely populated areas. 
These challenges become more severe in time-critical tasks, such as medical emergency response or disaster relief. 
For example, during a firefighting mission, a robot may need to transport emergency supplies through dense crowds of evacuees. The resulting high-density counter-flows can easily block the robot's path and delay the mission. Such scenarios highlight the need for navigation systems that can achieve both safety and agility in dynamic and {crowded} environments.

Navigation in these environments traditionally follows a perception-planning-control pipeline. Within this framework, classical planning methods for dynamic environments have been developed, including the Dynamic Window Approach (DWA) \cite{fox1997} and Velocity Obstacle (VO) methods \cite{fiorini1998} . 
By reasoning in the velocity domain instead of the spatial domain, these methods can impose explicit collision-avoidance constraints. However, they rely on strong assumptions, such as  accurate obstacle motion states (position and velocity) and environmental maps. 
Such assumptions are adopted in prior studies \cite{han2022, chen2024, qin2024, martinez-baselga2025}, where privileged obstacle states (i.e., ground truth position and velocity for each obstacle) are available. However, these assumptions seldom hold in practice. 
To relax these assumptions, some studies attempt to estimate the obstacle states using onboard detection and tracking (e.g., YOLO or MHT) from local observations \cite{xie2023, xu2025, huajian2024, xiao2025}. Nevertheless, in dense crowds, severe occlusion and sensor noise often cause unreliable state estimation, which in turn degrades the performance of Velocity Obstacle (VO)-based planners.

\begin{figure}[t]
    \centering
    \includegraphics[width=1\linewidth]{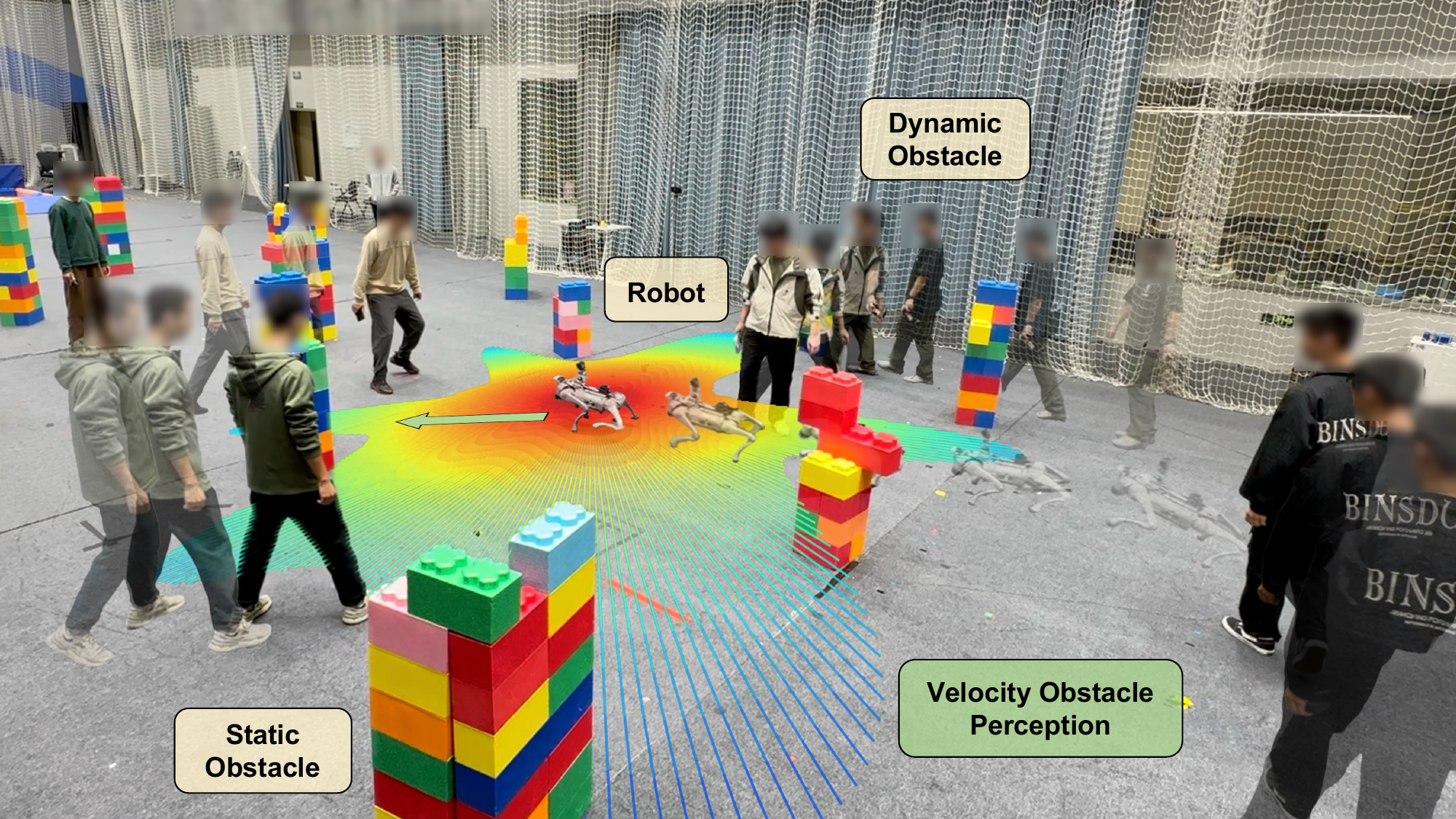}
   \caption{Real-world deployment of VOP-Nav on a Unitree Go2 robot. The overlay visualizes the safe velocity region predicted by VOP-Net, where the gradient from red to blue denotes collision risk from high to low.} 
    \label{fig:teaser}
\end{figure}

An alternative direction is end-to-end learning. By training policies via  reinforcement learning \cite{he2024} or differentiable physics \cite{zhang2024}, these approaches directly map sensory observations to control actions. This approach reduces the need for explicit perception and planning modules, and avoids error accumulation and delays caused by multi-stage pipelines. For quadruped robots, end-to-end systems that output joint-level commands can further enhance agility \cite{he2024}, which is critical for passing through transient free spaces in crowded environments.

Despite the strong results reported in \cite{he2024}, end-to-end approaches often show reduced performance in dense and dynamic environments, as observed in our experiments. This limitation may come from two closely related factors. {First, designing effective rewards for avoiding dense moving obstacles remains challenging. In such scenes, an effective avoidance reward should encourage the policy to account for changes in the surrounding environment rather than only react to imminent collisions. Conventional collision- or distance-based rewards can train reactive avoidance behaviors, but they provide limited guidance for avoiding crowded regions that may later become unsafe.}
{Second, end-to-end policies rely on implicit representations learned from raw sensor observations. In dense dynamic scenes, such representations may not reliably capture the time-varying collision constraints imposed by moving obstacles. Together, these limitations make it difficult for the policy to effectively avoid regions that may become unsafe in highly dynamic and crowded scenes.}

These observations highlight the need for end-to-end navigation frameworks that can explicitly consider obstacle motion, while retaining the benefits of direct sensor-to-action learning.
Previous studies  \cite{han2022, chen2024, qin2024, xie2023, xu2025, huajian2024} show that combining learning with Velocity Obstacle principles can improve navigation performance. However, most existing methods still rely on the standard Velocity Obstacle computations, which require accurate estimates of each obstacle's position and velocity. As discussed earlier, obtaining such information in crowded environments is difficult due to frequent occlusions and the challenge of detecting and tracking unknown or unseen object types (e.g., rarely observed animals).

To address these challenges in crowded navigation, we propose a novel end-to-end quadruped navigation system, \textbf{Velocity Obstacle Perception Navigation (VOP-Nav)}. 
Our system maps onboard depth camera and LiDAR observations directly to joint-level commands, combining the safety guarantees of velocity obstacle (VO) concepts with the agility of low-level control.
At the core of our system is the \textbf{Velocity Obstacle Perception Network (VOP-Net)}, which addresses the perception bottleneck of conventional VO methods. 
Unlike approaches that rely on explicit obstacle detection and motion estimation, VOP-Net implicitly encodes dynamic environmental constraints by processing multi-frame LiDAR data to directly predict the safe velocity region (see Fig. \ref{fig:teaser}). 
These predictions guide the navigation policy, serving both as state inputs during inference and as reward signals during training. 
This design removes the need for explicit obstacle state estimation or tracking, while enabling the robot to safely exploit transient gaps that existing methods would miss. 

We conduct extensive experiments comparing our method with state-of-the-art approaches in crowded dynamic environments. 
Results show that our method achieves higher navigation success rates while maintaining faster locomotion speeds. 
Ablation studies confirm that the proposed VOP-Net plays a key role in improving collision avoidance. 
Finally, the navigation policy, trained entirely in simulation, is deployed zero-shot on a Unitree Go2 quadruped robot, demonstrating reliable real-world operation using only onboard sensors, without fine-tuning or access to privileged obstacle information.  
The main contributions of this paper are summarized as follows:

\begin{enumerate}
\item \textbf{VOP-Nav}, a unified end-to-end navigation framework that integrates velocity obstacle (VO) principles with reinforcement learning, relying solely on local observations to achieve superior performance in dense dynamic environments, {without the need for map-based planning or explicit obstacle detection and tracking}.
\item \textbf{VOP-Net}, a perception network that incorporates VO concepts to predict the safe velocity region, providing robust perception in crowded environments without depending on unreliable detection or tracking.
\item  A comprehensive simulation framework built on Isaac Gym with diverse dense dynamic scenarios, enabling robust policy training and evaluation of generalization to unseen environments.
\end{enumerate}

\section{Related Work}
\subsection{Legged Locomotion and Navigation}
Locomotion and navigation are inherently coupled, yet classical paradigms treat them as two separate layers, with locomotion handled as low-level control and navigation as high-level planning.

Locomotion methods for legged robots, predominantly model-based, have demonstrated strong performance in structured or relatively flat terrains. However, their adaptability to complex environments remains limited, as control policies are often constrained by conservative safety margins\cite{bledt2018,dicarlo2018,grandia2023,kim2019,neunert2018}. More recently, reinforcement learning has enabled significant progress in agile legged locomotion through efficient simulation-based training\cite{todorov2012,makoviychuk2021}. These approaches have achieved impressive capabilities such as high-speed running\cite{binpeng2020,hwangbo2019,ji2022,margolis2024}, parkour\cite{cheng2024,zhuang2024,zhuang2023}, pose recovery\cite{lee2019,ma2023}, and even advanced skills such as playing soccer\cite{haarnoja2024}, skateboarding\cite{liu2025}, and whole-body control\cite{fu2023,liu2024}.

Navigation, on the other hand, conventionally follows {the} mapping-planning-control pipeline. Within this modular framework, planning modules typically employ probabilistic search algorithms, gradient-based trajectory optimization, or learning-based planners, while locomotion models serve as the control layer~\cite{macenski2020,marder-eppstein2010,zhou2019}.
{This conventional pipeline performs well in many scenarios, particularly for long-horizon tasks that require robust, global planning. However, when} applied to legged robots, these hierarchical frameworks often simplify the complex quadrupedal dynamics into a holonomic mobile base model. This simplification not only fails to exploit the robot's inherent agility but also introduces risks of latency accumulation and error propagation across modules.

End-to-end policies provide an alternative by directly mapping sensor observations to control actions. Zhang et al. demonstrated this paradigm through a differentiable physics framework for high-speed drone navigation\cite{zhang2024}. Similarly, He et al. proposed ABS\cite{he2024}, an agile collision-free navigation policy coordinated by a Reach-Avoid Value Network. Although ABS achieves reliable navigation in static environments through policy switching, our work draws inspiration from this end-to-end paradigm but targets highly dense dynamic environments. However, unlike ABS, we incorporate Velocity Obstacle principles into both the training and inference processes. This integration enhances the perception of dynamic obstacles and significantly improves system agility.

\begin{figure*}[t]
    \centering
    \includegraphics[width=1.0\textwidth]{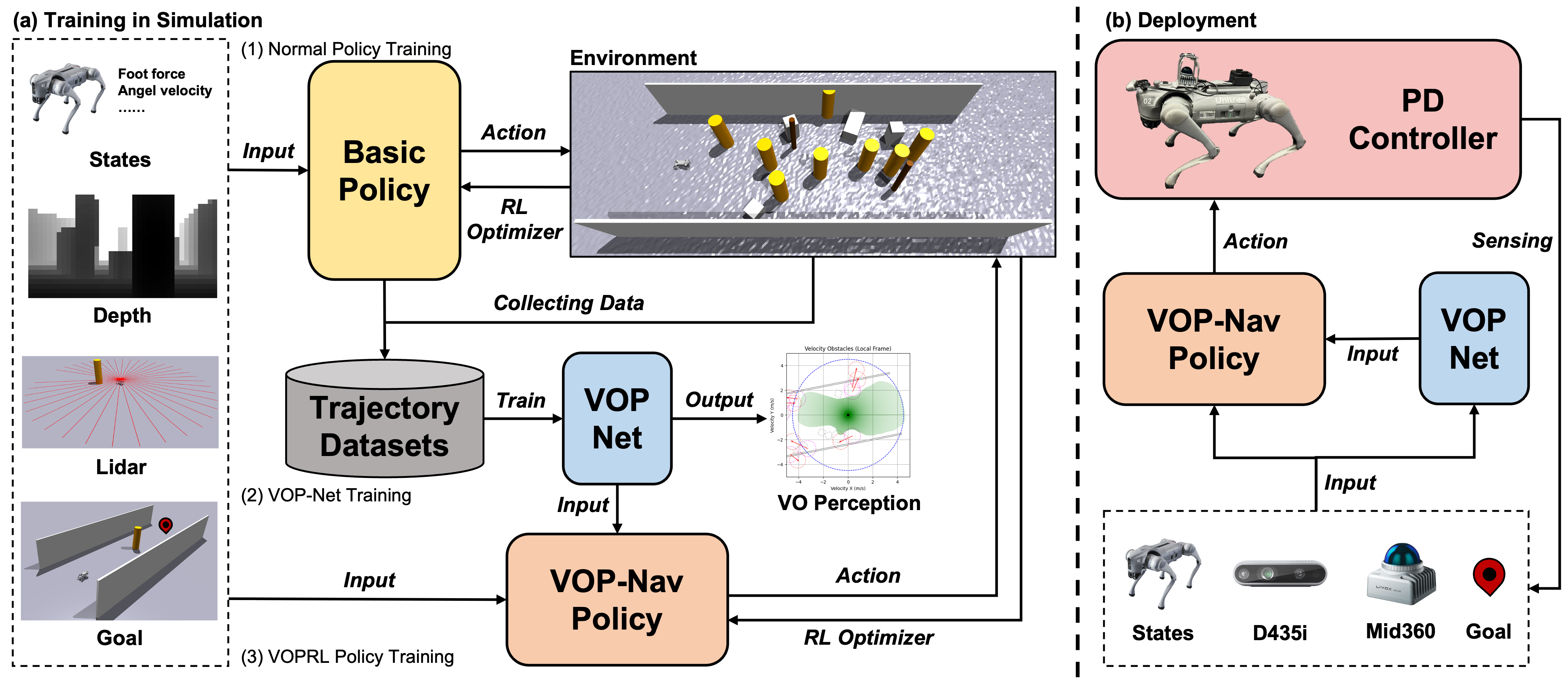}
    \caption{System Architecture: (a) Three-stage training pipeline in simulation: (1) Training a Basic policy in a dynamic environment; (2) Collecting trajectory data using the Basic policy to train VOP-Net; (3) Training the VOP-Nav Policy by incorporating VOP-Net predictions as both input and reward signal. (b) Real-world deployment on the Unitree Go2 quadruped robot. The VOP-Nav Policy and VOP-Net run concurrently to achieve real-time control, coupled with a 200 Hz low-level PD controller.}
    \label{fig:system}
\end{figure*}

\subsection{Navigation in Dynamic Crowded Environments}
Navigation in dynamic crowded environments involves multiple tightly coupled challenges, including collision avoidance, detection, tracking, and motion planning. Traditional approaches typically rely on explicit modeling of dynamic agents and trajectory prediction to enable safe navigation. Among these methods, the Social Force model~\cite{helbing1998} has been widely adopted to describe pedestrian interactions in crowds.

Velocity Obstacle (VO) methods form another cornerstone of dynamic collision avoidance by mapping obstacles into the velocity domain and defining collision cones\cite{fiorini1998}. Extensions such as Reciprocal Velocity Obstacle (RVO)\cite{vandenberg2008} and ORCA\cite{vandenberg2011} further address oscillation issues and enable reciprocal collision avoidance in multi-agent settings. Subsequent methods, including HRVO\cite{snape2011}, NH-ORCA\cite{alonso-mora2013}, and PRVO\cite{gopalakrishnan2017} expanded these concepts to hybrid and non-holonomic systems. Recent advancements such as AVOCADO \cite{martinez-baselga2025} and DGVO \cite{xiao2025} have further improved the Velocity Obstacle framework by addressing intent prediction for non-cooperative agents and modeling robotic motion constraints respectively.

To combine geometric safety with learning-based adaptability, recent studies have integrated VO principles into reinforcement learning. These VO-aided approaches can be categorized into two groups. The first group assumes access to privileged obstacle states to construct VO constraints. For example, RL-RVO\cite{han2022} encodes RVO features into the RL state space, while Chen et al.\cite{chen2024} employ graph attention and transformer architectures to extract VO-related representations, with similar geometric integrations explored in\cite{qin2024,zhou2022,zhou2023}. However, these methods need accurate states of both static and dynamic obstacles to compute VO features, which is rarely available in real-world deployment. 

The second group  estimates obstacle states using onboard perception. DRL-VO\cite{xie2023} and NavRL\cite{xu2025} use visual detectors (e.g., YOLO) to recover pedestrian states for VO construction. Although this removes external dependencies, performance degrades significantly in dense crowds due to frequent occlusions and detection failures. Moreover, such approaches are typically limited to predefined semantic categories (e.g., pedestrians), restricting their ability to generalize to diverse and previously unseen dynamic obstacles.

Beyond VO-based integrations, other learning approaches address crowded navigation through explicit interaction modeling or fully end-to-end perception. Interaction-centric methods explicitly model agent interactions, often using spatio-temporal graphs or attention mechanisms. For instance, HEIGHT proposed by Liu et al.\cite{liu2025a} represents robot–human and human–human interactions via a heterogeneous graph with dual attention, while related works follow similar formulations \cite{chen2019a,medina2023,zhou2023a}. These approaches generally depend on estimation of obstacle motion states, making them sensitive to perception and tracking reliability. 

Alternatively, end-to-end navigation methods directly map raw sensor observations to control actions without explicit state estimation \cite{fan2018,ahmed2022,pfeiffer2017,yuan2025reasan}. For example, Ahmed et al. \cite{ahmed2022} employ deep convolutional architectures to infer navigation commands directly from LiDAR or depth data. While such approaches simplify the perception pipeline, their fully implicit representations often struggle to capture the structured constraints of dynamic environments, leading to suboptimal performance in highly dynamic and crowded scenarios.

In summary, existing approaches either rely on explicit obstacle state estimation, which is fragile in crowded environments, or lack explicit safety-aware representations for dynamic collision avoidance. Our proposed VOP-Nav addresses this gap by directly perceiving the safe velocity region from historical LiDAR observations, combining the safety advantages of VO with the robustness of end-to-end learning, without relying on explicit detection or tracking modules.

\section{Method Overview}
\subsection{{Problem Formulation}}
{We formulate safe and agile collision-free navigation with onboard perception as a partially observable Markov decision process (POMDP), because the robot can only observe a partial and occlusion-prone view of the dynamic environment through onboard LiDAR, depth, and proprioceptive sensing. A POMDP can be formally defined as a 6-tuple: $\langle \mathcal{S}, \mathcal{A}, \mathcal{T}, \mathcal{R}, \Omega, \mathcal{O} \rangle$, where $\mathcal{S}$ denotes the environment state space, $\mathcal{A}$ represents the action space, $\mathcal{T}$ is the state transition function $T(s_{t+1} \mid s_t, a_t)$, $\mathcal{R}$ is the reward function, $\Omega$ is the onboard observation space, and $\mathcal{O}$ is the observation function $O(o_t \mid s_t)$.}

{At each time step $t$, the robot receives an observation $o_t \in \Omega$ and selects an action $a_t \in \mathcal{A}$ according to its policy $\pi_{\theta}(a_t \mid o_t)$. Upon executing $a_t$, the environment transitions from $s_t$ to $s_{t+1}$ with probability $T(s_{t+1} \mid s_t, a_t)$, generates the next observation $o_{t+1}$ with probability $O(o_{t+1} \mid s_{t+1})$, and provides reward $r_t = \mathcal{R}(s_t, a_t)$. The objective is to learn a policy $\pi_{\theta}$ that maximizes the expected discounted return}
\begin{equation}
{
J(\pi_{\theta}) = \mathbb{E}_{\pi_{\theta}}\left[ \sum_{t=0}^{\infty} \gamma^t r_t \right],
}
\label{eq:POMDP}
\end{equation}
{where $\gamma \in [0, 1)$ is the discount factor. We optimize $\pi_{\theta}$ with Proximal Policy Optimization (PPO)~\cite{schulman2017}.}

\subsection{{System Framework}}
The proposed system is illustrated in Fig.~\ref{fig:system} and adopts three training stages. In the first stage, a Basic policy is trained within a dynamic simulation environment (Isaac Gym) to acquire basic navigation capabilities and generate data for {the} next stage. The policy input includes the robot's proprioceptive state (e.g., foot contact forces, {angular velocities}, and projected gravity vectors), exteroceptive data (depth and LiDAR), and target goal information. Goal positions are randomly sampled at the start of each episode to encourage diverse behaviors.

In the second stage, the trained Basic policy is used to collect trajectory data for training the Velocity Obstacle Perception Network (VOP-Net). This dataset contains robot observations and states, alongside the geometric attributes and kinematic states (position and velocity) of surrounding dynamic and static obstacles. The supervision signal for VOP-Net, termed the VO-safety region, is computed from velocity obstacle theory and represented as a  $360^{\circ}$ safe velocity range. Details of the dataset construction and training procedure are provided in Section~\ref{sec:vop_net_vogt}. 

In the final stage, the navigation policy (VOP-Nav) is trained. Here, the state space is augmented with the VOP-Net output, combining it with the inputs from the first stage. Crucially, a specific reward function incorporating VOP-Net predictions is introduced during policy training to guide safe velocity selection, thereby reinforcing learning collision avoidance behaviors in dynamic environments.

For real-world validation, the trained VOP-Nav policy is deployed on a Unitree Go2 quadruped robot equipped with an onboard depth camera and a LiDAR. During deployment, the VOP-Net and VOP-Nav policy run in parallel to ensure real-time inference. The navigation policy is executed at 50~Hz, while a low-level PD controller runs at 200~Hz to maintain stable locomotion. Further implementation details and experimental settings are provided in Section~\ref{sec:exp}.

\section{Velocity Obstacle Perception Network (VOP-Net)}
The VOP-Net is designed to extract dynamic safety constraints directly from raw sensor data. We refer to this approach as ``Velocity Obstacle Perception''. 
Unlike traditional methods that explicitly compute Velocity Obstacle (VO) geometry from estimated obstacle states, VOP-Net implicitly captures the motion constraints of surrounding obstacles by predicting a safe velocity region. This design allows the network to represent essential dynamic environmental constraints without relying on accurate obstacle detection, tracking, or explicit VO computation.

\subsection{Velocity Obstacle}
\label{sec:system_vo}
\begin{figure}
    \centering
    \includegraphics[width=0.45\textwidth]{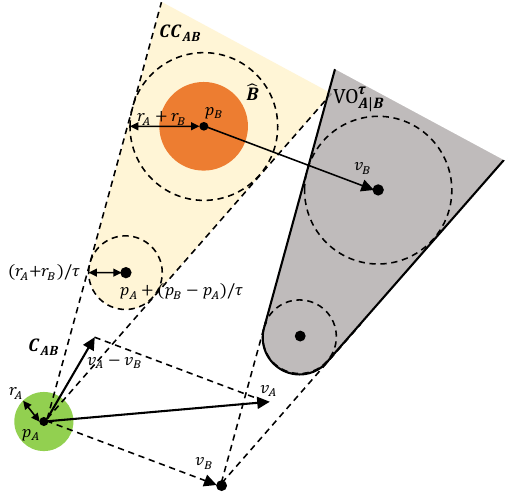}
    \caption{Geometric interpretation of the Velocity Obstacle formulation. The truncated collision cone $CC_{AB}$ (beige region) represents the set of relative velocities leading to a collision within time horizon $\tau$. This region is obtained by excluding the inner portion of the cone $C_{AB}$ where collisions would occur after $\tau$. The final Velocity Obstacle $\text{VO}^\tau_{A|B}$ (gray region) is constructed by translating $CC_{AB}$ by the obstacle's velocity $v_B$ (Minkowski sum). Any robot velocity $v_A$ falling within the gray region will result in a collision.}
    \label{fig:vop}
\end{figure}

The velocity obstacle formulation follows ORCA\cite{vandenberg2011}. We consider robot A and obstacle B in the velocity domain as shown in Fig. \ref{fig:vop}. To simplify the analysis, we model robot A as a circle with radius $r_A$, position $p_A$ and velocity $v_A$. Although obstacles can take various shapes, we assume obstacle B to be a circular obstacle with radius $r_B$, position $p_B$ and velocity $v_B$ for clarity. First, we define the original cone $C_{AB}$, which represents all potential collision velocities without temporal constraints: 
\begin{equation}
C_{AB} = \{v_{AB}|\lambda_{AB} \cap \hat{B} \neq \emptyset \}
\label{eq:C_AB}
\end{equation}
where $v_{AB} = v_A-v_B$ denotes the relative velocity of robot A with respect to obstacle B. $\lambda_{AB}$ represents the ray originating from $p_A$ in the direction of $v_{AB}$. $\hat{B}$ is the circle centered at $p_B$ with radius $(r_A + r_B)$. Geometrically, when we regard robot A as a point mass, $\hat{B}$ represents the expanded collision region. Accordingly, $C_{AB}$ can be interpreted as the cone of all relative velocity directions that could potentially lead to a collision if given sufficient time. However, in practical applications, we are only concerned with collisions that occur within a specific time horizon $\tau$. Therefore, we define the truncated collision cone $CC_{AB}$ by removing the inner portion of $C_{AB}$ that corresponds to collisions occurring after time $\tau$. The collision cone $CC_{AB}$ excludes velocity directions that correspond to collisions in more than $\tau$ time units, retaining only those that result in collision within time $\tau$, as shown in {Fig. \ref{fig:vop}}.

Let $\text{VO}^\tau_{A|B}$ denote the velocity obstacle domain, representing the set of velocities for robot A that would lead to a collision with obstacle B within the next $\tau$ time steps. The $\text{VO}^\tau_{A|B}$ can be represented as:

\begin{equation}
\text{VO}^\tau_{A|B} = \left\{ v_A  \middle|  v_A\in CC_{AB} \oplus v_B \right\}
\label{eq:VO}
\end{equation}
where $\oplus$ denotes the Minkowski sum, geometrically corresponding to translating the truncated collision cone by the velocity vector $v_B$. This translation transforms the relative velocity space to the absolute velocity space of robot A. A collision will occur within time $\tau$ if the velocity vector $v_A$ belongs to the set $\text{VO}^\tau_{A|B}$. Equivalently, one can say that a collision occurs when the relative velocity $v_A - v_B$ lies within $CC_{AB}$. For non-circular obstacles (e.g., rectangles or lines), we extract their corner points and treat them as circular obstacles with radius $r_A$ to apply the standard VO computation. The detailed calculation process is further described in {Section \ref{sec:vop_net_vogt}}.

\subsection{Safe Velocity Region}
\label{sec:vop_net_vogt}
\begin{figure*}[t]
    \centering
    \includegraphics[width=0.9\textwidth]{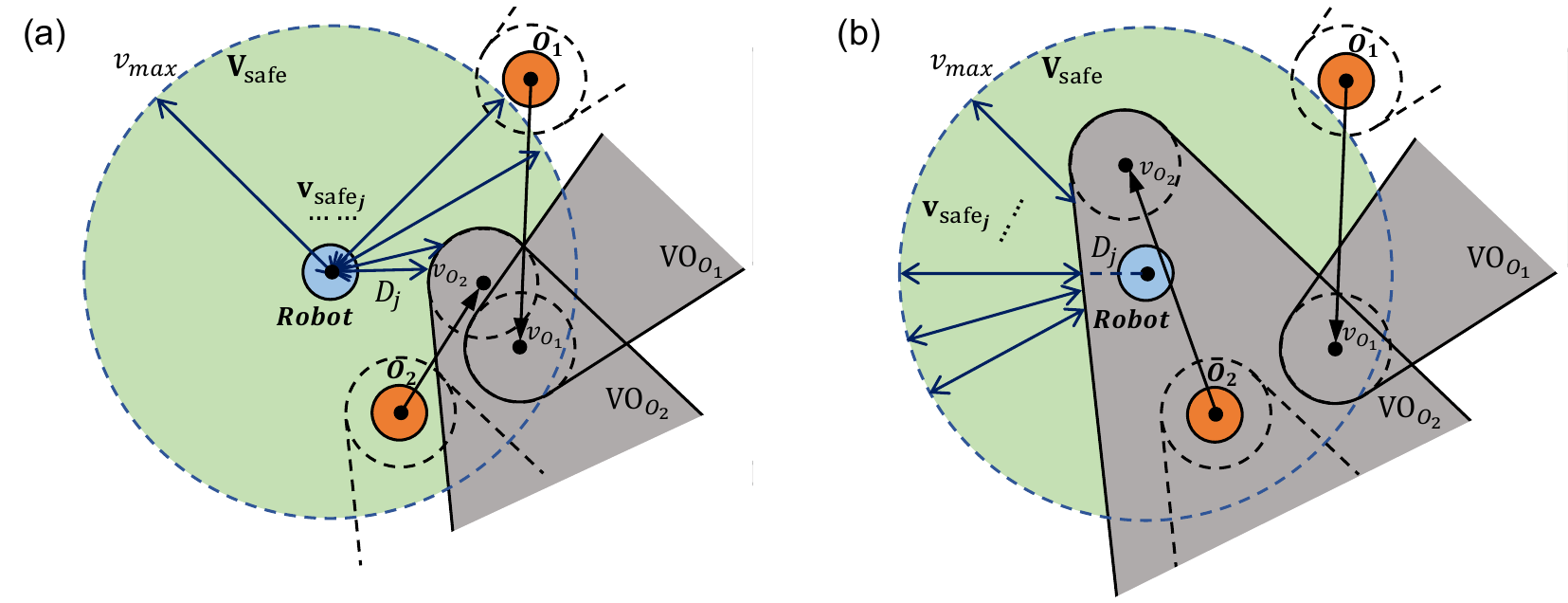} 
    \caption{Illustration of the Safe Velocity Region Computation. The robot (blue) and obstacles (orange) determine the kinematic constraints. The grey area represents the union of Velocity Obstacles ($\Omega_{\text{VO}}$), and the dashed circle indicates the robot's maximum speed ($v_{\text{max}}$). The safe velocity region ($V_{\text{safe}}$, green area) is the intersection of the feasible velocity set and the complement of $\Omega_{\text{VO}}$. The blue double-arrowed line represents the discretized safe velocity interval $\mathbf{v}_{\text{safe}_j}$ for a specific direction $j$, determined by critical velocity magnitude $D_j$. (a) Robot outside the VO region: safe interval $\mathbf{v}_{\text{safe}_j} = [0, D_j]$. (b) Robot inside the VO region: safe interval $\mathbf{v}_{\text{safe}_j} = [D_j, v_{\text{max}}]$.}
    \label{fig:safevelcompute}
\end{figure*}

To train the VOP-Net network, we need to first compute the safe velocity region as the supervision signal. At each time step, the safe velocity region is defined as the intersection of the robot's maximum feasible velocity set and the complement of the union of VO regions generated by all surrounding obstacles, as shown in Fig.~\ref{fig:safevelcompute}. However, in crowded scenes, the union of VO regions often has a complex, irregular geometry that is difficult for network training. Therefore, we discretize the safe velocity region over the angular domain (spanning $360^{\circ}$). This transformation maintains clear physical interpretability while providing a significantly more structured and learnable training target for the neural network.

Specifically, we transform obstacle states from the global world coordinate frame $F_W$ to the robot's local coordinate frame $F_R$. This transformation expresses motion constraints in the robot's egocentric reference frame, facilitating the selection of safe velocities. Let the positions of $n$ obstacles be denoted as $p_{O_i} = [x_{O_i}, y_{O_i}, z_{O_i}]^T \in \mathbb{R}^3$ for $i=1,2,\ldots,n$. Since the navigation task primarily concerns planar motion, we focus on the projection onto the XY-plane. The affine transformation from the robot frame $F_R$ to the world frame $F_W$, denoted as $T_{R \rightarrow W}$, is determined by the robot's current position $p_R = [x_R, y_R, z_R]^T$ and heading $\theta_R \in [-\pi, \pi)$. Constructing the transformation matrix using the planar components:

\begin{equation}
T_{R \rightarrow W} =
\begin{bmatrix}
R(\theta_R) & t \\
0^T & 1
\end{bmatrix}
\end{equation}
where $t = [x_R, y_R]^T$ is the translation vector and $R(\theta_R)$ is the rotation matrix:
\begin{equation}
R(\theta_R) =
\begin{bmatrix}
\cos\theta_R & -\sin\theta_R \\
\sin\theta_R & \cos\theta_R
\end{bmatrix}.
\end{equation}

Subsequently, the inverse transformation $T_{W \rightarrow R} = T_{R \rightarrow W}^{-1}$ is computed to map obstacle states into the robot's local frame. Let $\tilde{p}_{O_i} = [x_{O_i}, y_{O_i}, 1]^T$ represent the homogeneous planar coordinates of obstacle $i$. The local position $p'_{O_i}$ is obtained by calculating the transformed homogeneous coordinates $\tilde{p}'_{O_i} = T_{W \rightarrow R} \,\, \tilde{p}_{O_i}$ and extracting the first two components. Similarly, the obstacle velocities are transformed using the rotation component: $v'_{O_i} = R^T(\theta_R) v_{O_i}$.

Using the transformed local coordinates, we compute the individual velocity obstacle $\text{VO}_{O_i}$ for each entity as described in Section~\ref{sec:system_vo} and denote their union as $\Omega_{\text{VO}} = \bigcup_{i=1}^{n} \text{VO}_{O_i}$. Assuming a kinematic limit $v_{\text{max}}$, the set of max feasible velocities forms a circle denoted as $V_{\text{max}}$. Consequently, the safe velocity region is naturally defined as the intersection of the feasible velocity set with the complement of the collision region: $V_{\text{safe}} = V_{\text{max}} \cap \, \overline{\Omega}_\text{VO}$. We uniformly discretize the angular space into $d$ directions. Accordingly, we define the discretized safe velocity region $\mathbf{V}_{\text{safe}}$ as a set of directional safe velocity intervals: $\mathbf{V}_{\text{safe}} = \{ \mathbf{v}_{\text{safe}_1}, \mathbf{v}_{\text{safe}_2}, \ldots, \mathbf{v}_{\text{safe}_d} \}$, where each $\mathbf{v}_{\text{safe}_j}$ represents the feasible velocity range along the $j$-th direction. Due to the geometric nature of VO, disjoint safe segments may exist along a single direction. {For a compact representation that is easy to express mathematically and suitable for fixed-size network regression, when multiple safe segments appear along one ray, we retain a single continuous safe interval as the effective interval. {Specifically, we select the lower-speed feasible interval along each direction as a conservative safety-oriented design. Although higher-speed intervals satisfy the theoretical maximum-speed bound, the learned locomotion policy may not reliably achieve the corresponding velocities. Encouraging the navigation policy to rely on these intervals may also induce overly aggressive behaviors.} Following this discretization, the discretized safe velocity region is represented by the collection of $d$ directional safe velocity intervals. This discretized safe velocity region is then used as the supervision signal for training VOP-Net}.

\begin{algorithm}[ht]
\caption{Safe Velocity Region Computation}
\label{alg:1}
\begin{algorithmic}[1]
\REQUIRE Robot position $p_R = [x_R, y_R, z_R]$, yaw angle $\theta_R$, obstacle states $\{p_{O_i}, v_{O_i}\}_{i=1}^n$, number of directions $d$, maximum speed $v_{\text{max}}$
\ENSURE safe velocity region $\mathbf{V}_{\text{safe}}$
\FOR{each obstacle $O_i$}
    \STATE Transform position: $\tilde{p}'_{O_i} = T_{W \rightarrow R} \cdot \tilde{p}_{O_i}$, extract first two components to get $p'_{O_i}$
    \STATE Transform velocity: $v'_{O_i} = R^T(\theta_R) \cdot v_{O_i}$
    \FOR{each direction $\lambda_j$, $j = 1, \dots, d$}

        \STATE Compute $L_j \leftarrow$ distance from origin to Minkowski boundary $\hat{B}$ and tangents.

        \STATE $D_j \leftarrow \min(L_j, v_{\text{max}})$
        \IF{robot outside VO region}
            \STATE $\mathbf{v}_{\text{safe}_j}^{O_i} \leftarrow [0, D_j]$
        \ELSE
            \STATE $\mathbf{v}_{\text{safe}_j}^{O_i} \leftarrow [D_j, v_{\text{max}}]$
        \ENDIF
    \ENDFOR
    \STATE $\mathbf{V}_{\text{safe}}^{O_i} \leftarrow \{\mathbf{v}_{\text{safe}_j}^{O_i}\}_{j=1}^d$
\ENDFOR
\STATE $\mathbf{V}_{\text{safe}} \leftarrow \bigcap_{i=1}^n \mathbf{V}_{\text{safe}}^{O_i}$
\RETURN $\mathbf{V}_{\text{safe}}$
\end{algorithmic}
\end{algorithm}

In practice, the computation of the safe velocity region $\mathbf{V}_{\text{safe}}$ involves determining and intersecting the feasible velocity intervals imposed by each individual obstacle $O_i$. This procedure is summarized in Algorithm~\ref{alg:1}. The process initiates by updating all obstacle states relative to the robot's local frame. Subsequently, for each discrete direction $\lambda_j$, we determine the boundary distance $L_j$ from the robot's origin to the expanded obstacle region $\hat{B}$, where  $\hat{B}$ corresponds to the standard Minkowski sum boundary.

Based on the boundary distance $L_j$, the safe velocity interval along direction $\lambda_j$ is determined by identifying the feasible segment closest to the origin. Specifically, we define the critical velocity magnitude as $D_j = \min(L_j, v_{\text{max}})$. As illustrated in Fig.~\ref{fig:safevelcompute}, if the robot's current state lies outside the VO region, the feasible range is defined as $\mathbf{v}_{\text{safe}_j}^{O_i} = [0, D_j]$.  Conversely, if the robot is inside the VO region (implying an imminent collision risk), the system prioritizes the minimum velocity required to exit the collision state, yielding $\mathbf{v}_{\text{safe}_j}^{O_i} = [D_j, v_{\text{max}}]$. Aggregating these intervals across all obstacles yields the final safe velocity region:

\begin{equation}
\mathbf{V}_{\text{safe}} = \bigcap_{i=1}^n \mathbf{V}_{\text{safe}}^{O_i}, \quad \text{where} \quad \mathbf{V}_{\text{safe}}^{O_i} = \left\{\mathbf{v}_{\text{safe}_j}^{O_i}\right\}_{j=1}^d.
\end{equation}

\noindent This intersection is performed element-wise across the $d$ discrete directions to ensure that the resulting velocity command satisfies the safety constraints of the entire environment simultaneously.

\subsection{Network Architecture of VOP-Net}
\begin{figure*}[t]
    \centering
    \includegraphics[width=1\textwidth]{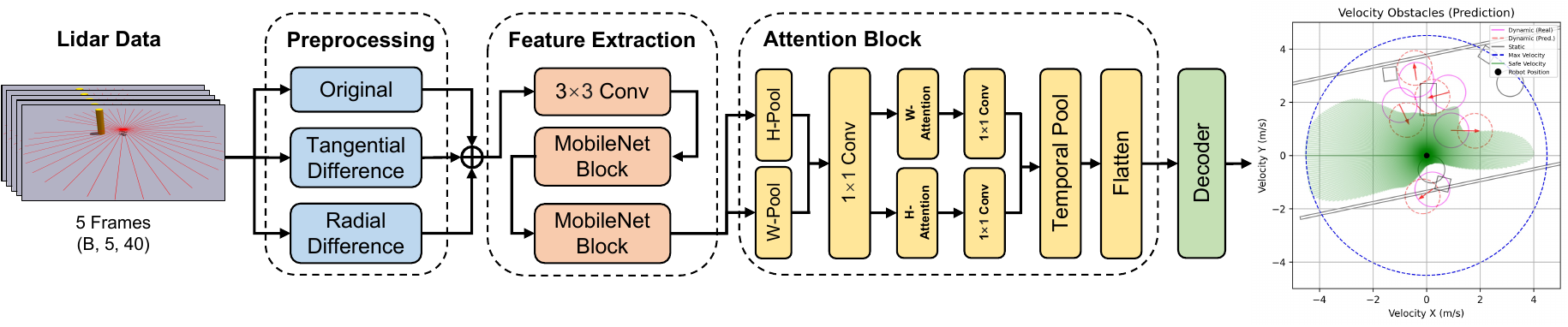}
    \caption{The architecture of VOP-Net. The pipeline begins with five-frame LiDAR scans. In the Preprocessing stage, the raw input is converted into a 3-channel tensor to explicitly encode distance, spatial, and temporal dynamics. Deep features are then captured with MobileNet blocks, followed by an Attention Block that re-weights features along orthogonal temporal and angular axes. Finally, the Decoder regresses these features to generate the output, representing the 360-degree safe velocity region (in green shadow).}
\label{fig:vopnet_arch}
\end{figure*}

The proposed VOP-Net architecture, as illustrated in Fig.~\ref{fig:vopnet_arch}, is designed to efficiently capture high-level spatiotemporal features from raw LiDAR sequences. The input consists of a history of LiDAR scans spanning the current frame and the preceding four frames. To explicitly encode environmental dynamics, we preprocess this input into a three-channel representation: the first channel contains raw distance measurements; the second encodes tangential differences (spatial gradients) to highlight obstacle boundaries; and the third {contains frame-to-frame radial differences (temporal gradients) that encode observation-level cues of relative displacement between the robot and its surroundings}. These preprocessed features are extracted with MobileNet blocks, providing a lightweight yet effective backbone for feature learning.

To further refine the representation, we employ an orthogonal attention mechanism that decomposes feature aggregation into two distinct directions. By performing pooling operations separately along the temporal axis and the spatial angular axis, the network generates independent attention weights for the time sequence and the directional sectors. This allows the model to simultaneously emphasize critical historical moments and hazardous spatial directions. Finally, the attended features are aggregated via temporal pooling and mapped through an MLP decoder to regress the $d \times 2$ safe velocity region (where $d=360$), forming the predicted safe velocity region $\mathbf{V}_{\text{pred}}$.

\section{Training Stage}
The proposed VOP-Nav Policy is designed to combine the agility of end-to-end learning with the safety of traditional Velocity Obstacle methods. To achieve this, we implement a three-stage training architecture. In Stage 1, we train a Basic Policy capable of high-speed locomotion and fundamental collision avoidance. In Stage 2, we deploy this policy to collect interaction data within the simulation environment to train the VOP-Net. Finally, in Stage 3, we integrate the VOP-Net predictions into the policy's observation space and incorporate them into the reward function to constrain velocity selection. The detailed training process is described below.

\subsection{Stage 1: Basic Policy Training}
In the first stage, we train a Basic policy following the agile policy method in ABS \cite{he2024}. Unlike ABS, we incorporate depth camera observations and extend the LiDAR perception range to 360 degrees. The purpose of the Basic policy is to equip the robot with fundamental locomotion and obstacle avoidance capabilities, which facilitates data collection for VOP-Net in the subsequent stage.

\subsubsection{Observation and Action Space}
The observation space includes both proprioceptive and exteroceptive data. {The 50-dimensional proprioceptive observation consists of} foot contact forces, base angular velocities, projected gravity vector, relative distance to the goal, remaining episode time, joint positions, joint velocities, and the action from the previous timestep. For exteroceptive perception, we use a depth camera to detect frontal obstacles. To ensure real-time performance, the resolution is set to $22 \times 40$ pixels. Additionally, a LiDAR is used to measure obstacle distances across a 360-degree range, represented by a 40-dimensional vector with a 9-degree angular resolution. {The Basic policy input therefore has 970 dimensions in total.}

The action space consists of 12-dimensional target joint positions. These targets are tracked by a PD controller. The control torques ($\tau$) are calculated as:
\begin{equation}
\label{eq:pd_controller}
\tau = K_p (t_a - q) - K_d \dot{q}
\end{equation}
where $t_a$ represents the target joint positions, $q$ and $\dot{q}$ are the current joint positions and velocities, respectively, and $K_p$ and $K_d$ are the proportional and derivative gains.

\subsubsection{Reward Function}
The reward function is primarily adapted from the ABS framework~\cite{he2024}, comprising standard components such as collision penalties, goal-tracking rewards, and regularization terms for joint torques and action smoothness. These terms ensure physically plausible and stable locomotion. The key element enabling high-speed navigation is the agility reward, $r_{\text{agile}}$, which is formulated as:

\begin{equation}
r_{\text{agile}} = \max \left\{ \text{ReLU} \left( \frac{v_x}{v_{\text{max}}} \right) \cdot \mathbf{1}(\text{dir}_{\text{good}}), \mathbf{1}(d_{\text{goal}} < \sigma_{\text{tight}}) \right\},
\end{equation}

\noindent where $v_{\text{max}}=4.5\,\text{m/s}$ is the maximum speed. The indicator $\mathbf{1}(\text{dir}_{\text{good}})$ is active when the heading error is within $105^{\circ}$, ensuring directional compliance. The second term $\mathbf{1}(d_{\text{goal}} < \sigma_{\text{tight}})$ maintains the reward value when the robot enters the goal proximity ($\sigma_{\text{tight}} = 0.5\,\text{m}$). This formulation of $r_{\text{agile}}$ guarantees the robust locomotion capability of the Basic policy.

\subsubsection{Network Architecture and Training}
To efficiently process heterogeneous sensor data, we design a hybrid encoder architecture optimized for real-time deployment. For visual perception, depth images are encoded by a lightweight CNN that integrates spatial feature compression with a channel mixer module. This design effectively extracts spatial patterns and facilitates cross-channel interaction, compensating for the limited resolution of the depth input. Simultaneously, LiDAR measurements are processed via dedicated MLPs to encode geometric obstacle information. These exteroceptive features are concatenated with the robot's proprioceptive state and fused through a two-layer residual block with LayerNorm, followed by an MLP decoder to generate the final action output.

We observed that Isaac Gym exhibits slow performance when rendering depth images in highly parallelized scenarios and lacks native LiDAR sensor support. To address these limitations, we employed ray casting to simulate both depth and LiDAR sensors simultaneously. Although this method introduces some distortion in the resulting depth images, the performance impact remains negligible given the inherently low resolution of our depth data. This approach achieves significant acceleration in simulation speed.

The policy is optimized using the Proximal Policy Optimization (PPO) algorithm~\cite{schulman2017} within an Actor-Critic framework. Both the actor and critic networks utilize the aforementioned architecture to output 12-dimensional joint targets and scalar value estimates, respectively. Training is conducted in Isaac Gym across 1,024 parallel environments.

\subsection{Stage 2: VOP-Net Training}
\subsubsection{Data collection}
To train VOP-Net, we constructed a large-scale dataset by collecting trajectories from the Basic policy navigating in dynamic simulation environments. The dataset comprises 10,000 data files, where each file aggregates transitions from 1,024 parallel environments over a horizon of 48 time steps. These trajectories encompass the essential state information, including robot kinematics and obstacle states, required to compute the supervision signal of safe velocity region $\mathbf{V}_{\text{safe}}$ via Algorithm~\ref{alg:1}. During data generation, the robot's maximum speed was set to $v_{\text{max}} = 4.5 \, \text{m/s}$, a deliberately high value chosen to encourage the exploration of high-speed maneuvers within physical limits.

\subsubsection{Loss Function}
For the training objective, we employ a hybrid loss function that synergizes the geometric alignment properties of Intersection-over-Union (IoU) with the gradient stability of Mean Squared Error (MSE). The total loss $\mathcal{L}$ is defined as a weighted sum:

\begin{equation}
    \mathcal{L} = \alpha \mathcal{L}_{\text{IoU}} + (1 - \alpha) \mathcal{L}_{\text{MSE}}
\end{equation}

\noindent where $\alpha$ is a balancing coefficient (set to 0.2). Let $\mathbf{v}_{\text{pred}_j} = [v_{\text{pred}_j}^{\min}, v_{\text{pred}_j}^{\max}]$ and $\mathbf{v}_{\text{safe}_j} = [v_{\text{safe}_j}^{\min}, v_{\text{safe}_j}^{\max}]$ denote the predicted and ground truth velocity intervals for the $j$-th direction, respectively. The IoU loss quantifies the geometric overlap:

\begin{equation}
    \mathcal{L}_{\text{IoU}} = \frac{1}{d} \sum_{j=1}^{d} \left( 1 - \frac{|\mathbf{v}_{\text{pred}_j} \cap \mathbf{v}_{\text{safe}_j}|}{|\mathbf{v}_{\text{pred}_j} \cup \mathbf{v}_{\text{safe}_j}| + \epsilon} \right)
\end{equation}

\noindent Simultaneously, the MSE loss ensures numerical precision for the interval boundaries:

\begin{equation}
    \mathcal{L}_{\text{MSE}} = \frac{1}{2d} \sum_{j=1}^{d} \left( (v_{\text{pred}_j}^{\min} - v_{\text{safe}_j}^{\min})^2 + (v_{\text{pred}_j}^{\max} - v_{\text{safe}_j}^{\max})^2 \right)
\end{equation}

\noindent By combining these terms, the network learns to capture the overall safe velocity region structure via IoU while refining the exact boundary values via MSE, leading to robust performance in dynamic environments.

\subsection{Stage 3: Navigation Policy (VOP-Nav) Training}
The VOP-Net predicts a safe velocity region over all directions. While the prediction could theoretically be integrated into the navigation process as strict geometric constraints—similar to traditional velocity obstacle methods—doing so is often suboptimal. Since the network's predictions are learned approximations from raw sensor data, they may contain biases or uncertainties. Enforcing them as strict constraints can lead to overly conservative behaviors or infeasible motions. 

Instead, we leverage these predictions to \textit{guide} the policy toward inherently safe velocity selection, rather than using them for post-hoc velocity modification. To this end, we propose the VOP-Nav Policy, which integrates VOP-Net predictions in two complementary ways: (i) as an augmented component of the policy state during inference, and (ii) as an extra term in the reward function during training. This dual integration allows the policy to develop an intrinsic understanding of dynamic safety constraints during training, resulting in navigation behaviors that are both robust and efficient in complex dynamic environments. {The VOP-Nav policy is trained from scratch in this stage.}

\subsubsection{Observation Space}
The observation space largely follows that of the Basic policy, with the addition of the robot's linear velocity and the output of VOP-Net. The VOP-Net generates a $360 \times 2$-dimensional vector representing safe velocity region (minimum and maximum speeds) across 360 uniformly distributed angles. This output is flattened into a 720-dimensional vector and integrated into the policy input. {With the additional 3-dimensional robot linear velocity and the 720-dimensional VOP-Net output, the VOP-Nav policy input has 1693 dimensions, compared with 970 dimensions for the Basic policy.}

\subsubsection{Reward Function}
To explicitly guide the policy toward compliance with the dynamic environmental constraints inferred by VOP-Net, we introduce a specific velocity constraint reward, denoted as $r_{\text{VO}}$. This term penalizes velocity commands that violate the predicted safe constraints, serving as a soft constraint during optimization. The reward is formulated as:

\begin{equation}
r_{\text{VO}} =
\begin{cases}
v - v_{s\text{-min}}, & \text{if } v < v_{s\text{-min}} \\
v_{s\text{-max}} - v, & \text{if } v > v_{s\text{-max}} \\
0, & \text{otherwise}
\end{cases}
\end{equation}

\noindent where $v = \|\mathbf{v}_{xy}\|$ represents the magnitude of the robot's current planar velocity. The terms $v_{s\text{-min}}$ and $v_{s\text{-max}}$ correspond to the minimum and maximum safe velocity limits for the current direction of motion.

Since VOP-Net outputs discrete predictions across finite angular bins, obtaining precise constraints for the robot's continuous velocity direction $\theta = {\operatorname{atan2}(v_y, v_x)}$ requires smooth mapping. We employ a local interpolation mechanism to retrieve specific bounds for $\theta$. Specifically, we identify the $k$ nearest angular sectors relative to $\theta$ and compute the effective $v_{s\text{-min}}$ and $v_{s\text{-max}}$ using inverse-distance weighting. {Here, we set $k=5$. Let $\lambda_i$ denote the angular direction of the $i$-th selected sector. The angular distance between $\theta$ and $\lambda_i$ is computed with periodic wrap-around:}
\begin{equation}
{\delta_i=\min\left(|\theta-\lambda_i|,\,2\pi-|\theta-\lambda_i|\right),
\quad i=0,\ldots,k-1}
\end{equation}
{The inverse-distance interpolation weight is computed as:}
\begin{equation}
{w_i =
\frac{(\delta_i+\epsilon)^{-1}}
{\sum_{j=0}^{k-1}(\delta_j+\epsilon)^{-1}},
\quad i=0,\ldots,k-1}
\end{equation}
{where $\epsilon$ is a small constant for numerical stability. Let $v_{s\text{-min},i}$ and $v_{s\text{-max},i}$ denote the predicted safe-speed bounds of the $i$-th selected sector. The effective bounds used in $r_{\text{VO}}$ are then obtained by weighted averaging:}
\begin{equation}
{v_{s\text{-min}}(\theta)=
\sum_{i=0}^{k-1}w_i v_{s\text{-min},i},
\quad
v_{s\text{-max}}(\theta)=
\sum_{i=0}^{k-1}w_i v_{s\text{-max},i}}
\end{equation}
This process {makes the reward signal smoother} and correctly handles the periodicity of the angular space (i.e., the wrap-around at $2\pi$). By penalizing deviations from these interpolated allowable ranges, the policy develops an intrinsic preference for collision-free velocities.

\subsubsection{Network Architecture and Training}
The VOP-Nav policy inherits the hybrid backbone architecture of the Basic Policy, maintaining the same processing modules for depth images, LiDAR, and proprioceptive states. The primary structural enhancement is the introduction of a dedicated MLP encoder designed to process the explicit safety constraints. Specifically, the 720-dimensional safe velocity vectors predicted by VOP-Net are encoded by this MLP into latent safety features. These features are then concatenated with the visual, lidar, and proprioceptive embeddings.

The training process follows the same PPO-based Actor-Critic framework employed in Stage 1. Crucially, to prevent the introduction of the safety reward ($r_{\text{VO}}$) from inducing overly conservative behavior or compromising locomotion speed, we increase the weighting coefficient of the agility reward $r_{\text{agile}}$. This rebalancing ensures that the policy maintains high-speed capabilities while learning to adhere to the dynamic safety boundaries.

\subsection{Training Environments in Simulation}
\begin{figure}[t]
    \centering
    \includegraphics[width=1.0\linewidth]{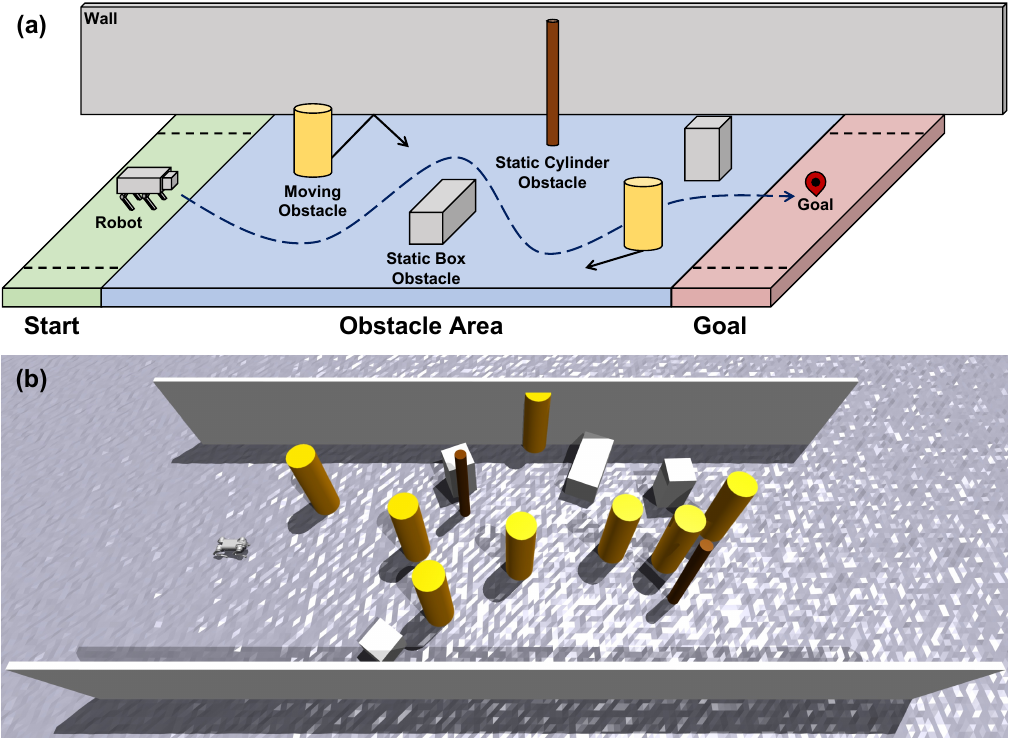}
    \caption{Training Environment. (a) Schematic: The robot navigates from the green start region to the pink goal region, targeting the specific red marker. The central zone contains linear-moving dynamic obstacles (yellow cylinders) and static obstacles (brown/gray), bounded by longitudinal walls. (b) Simulation View: The corresponding high-density scenario realization in Isaac Gym.}
    \label{fig:sim}
\end{figure}

As shown in Fig.~\ref{fig:sim}, we constructed a dynamic simulation training environment with Isaac Gym:

\subsubsection{Training Area}
Training is conducted in a $10 \times 5\,\text{m}^2$ rectangular arena. The start and goal regions, each measuring $0.5 \times 4 \,\text{m}^2$, are placed at opposite ends along the longitudinal axis, with start and target positions uniformly sampled within these regions. 
An $8 \times 5\,\text{m}^2$ central obstacle zone lies between them and is separated from the start and goal regions by safety buffers to avoid immediate collisions at initialization.
Boundary walls are installed along the longitudinal edges to prevent trivial solutions, such as bypassing the obstacle field through open side regions. This constrained layout forces the agent to traverse the dynamic obstacle zone, rather than relying on simple detours.
To improve computational efficiency, we model obstacles as 3D geometric primitives (cylinders and cuboids) rather than high-fidelity meshes. These primitives serve as effective collision proxies while minimizing simulation overhead.

To approximate the volumetric footprint of pedestrians, dynamic obstacles are modeled as cylinders ($h=1.7\,\text{m}, r=0.5\,\text{m}$), with up to eight instances per episode. These dynamic obstacles follow linear trajectories and undergo specular reflection at the arena boundaries to maintain continuous motion within the environment. Static elements represent structural diversity through two primary geometry types. Vertical objects are modeled as cylinders ($h=2.0\,\text{m}, r=0.2\,\text{m}$), with a maximum of two instances. Volumetric objects are represented by up to four cuboids, each randomly selected from three standardized sizes: $0.5 \times 0.5 \times 1.0$, $0.5 \times 0.5 \times 0.5$, and $1.2 \times 0.6 \times 0.6\,\text{m}^3$.

\subsubsection{Curriculum}
Adopting the adaptive curriculum strategy proposed by Zhang et al.~\cite{zhang2024a}, we implement a progressive training schedule that modulates task complexity based on agent performance. The curriculum level, denoted as $l_c \in \{0, 1, \ldots, 9\}$, is dynamically adjusted: it increments when the robot-to-goal distance falls below a promotion threshold ($0.5\,\text{m}$) and decrements if the distance exceeds a demotion threshold ($2.0\,\text{m}$); otherwise, it remains unchanged. The environmental parameters evolve according to $l_c$ as follows:

\begin{itemize}
    \item Obstacle Density: The number of obstacles scales linearly with the curriculum level, calculated as $n_o = \lfloor N_o \cdot 0.1 \cdot (1 + l_c) \rfloor$, where $N_o$ represents the maximum obstacle capacity.

    \item Velocity Magnitude: The velocity envelope for dynamic obstacles expands as complexity increases. The velocity magnitude is sampled from $[v^d_{\text{min}}, v^d_{\text{max}}]$, where the upper bound is governed by:
    \begin{equation}
    v^d_{\text{max}} = \max\left( v_{\text{min}}^{\text{obj}}, 0.1 \cdot (l_c + 1) \cdot v_{\text{max}}^{\text{obj}} \right)
    \end{equation}
    Here, $v_{\text{min}}^{\text{obj}}$ and $v_{\text{max}}^{\text{obj}}$ denote the global minimum and maximum velocity limits. The lower bound is constrained to $v^d_{\text{min}} = 0.5 \cdot v^d_{\text{max}}$ to ensure dynamic obstacles maintain a challenging level of mobility. Directions are initialized uniformly at random.

    \item Stochastic Behavior: To simulate unpredictable motion, dynamic obstacles are subjected to stochastic velocity updates every 100 simulation steps. We introduce a perturbation probability $p_{\text{change}} = 0.05 \cdot l_c$ that increases with difficulty. When a perturbation occurs (indicated by indicator $f_c \sim \text{Bernoulli}(p_{\text{change}})$), the obstacle either adjusts its acceleration (with probability 0.9, $f_a=1$) or halts (with probability 0.1, $f_a=0$). The acceleration magnitude is defined as $a_d = (r - 0.5) \cdot v_{\text{max}}^{\text{obj}} \cdot 0.1 \cdot (1 + l_c)$, where $r \sim \text{Uniform}(0, 1)$. The velocity update rule is formulated as:
    \begin{equation}
    v^d_{\text{new}} = v^d_{\text{old}} + \left( a_d \cdot f_a - (1-f_a) \cdot v^d_{\text{old}} \right) \cdot f_c
    \end{equation}
    Stopped obstacles resume motion with a randomly assigned velocity after a 100-step interval.

    \item Terrain Complexity: The environmental roughness is progressively augmented. As $l_c$ increases, the vertical amplitude of rough terrain and stumbling blocks scales linearly from $0\,\text{cm}$ to $7\,\text{cm}$.
\end{itemize}

\section{Experiments}
\label{sec:exp}
\begin{figure*}[t]
    \centering
    \subfloat[Forest Environment]{\includegraphics[width=0.24\textwidth]{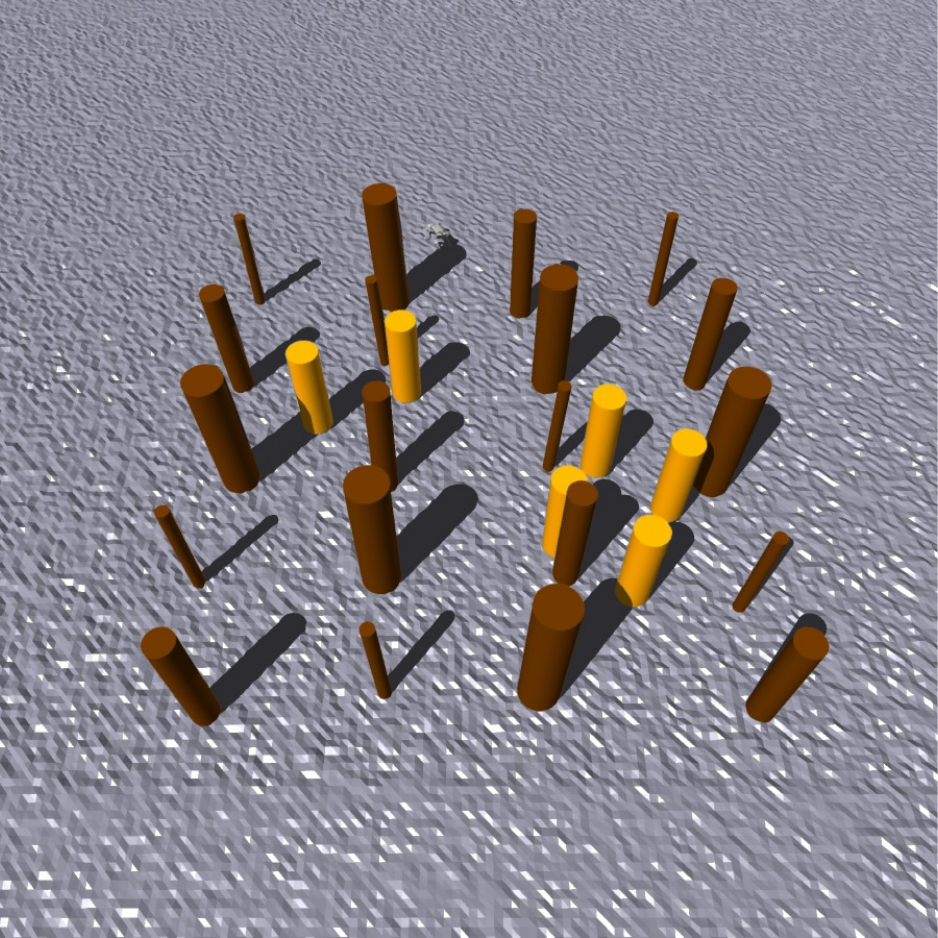}}
    \hfill
    \subfloat[Office Environment]{\includegraphics[width=0.24\textwidth]{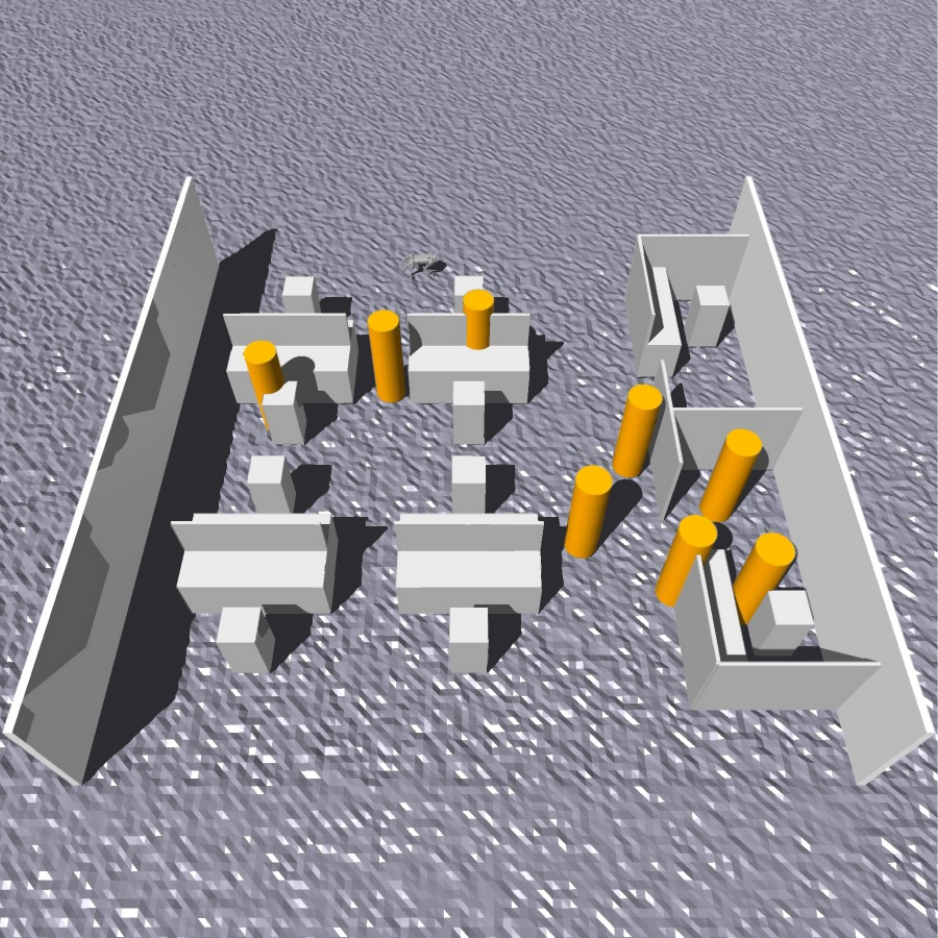}}
    \hfill
    \subfloat[Square Environment]{\includegraphics[width=0.24\textwidth]{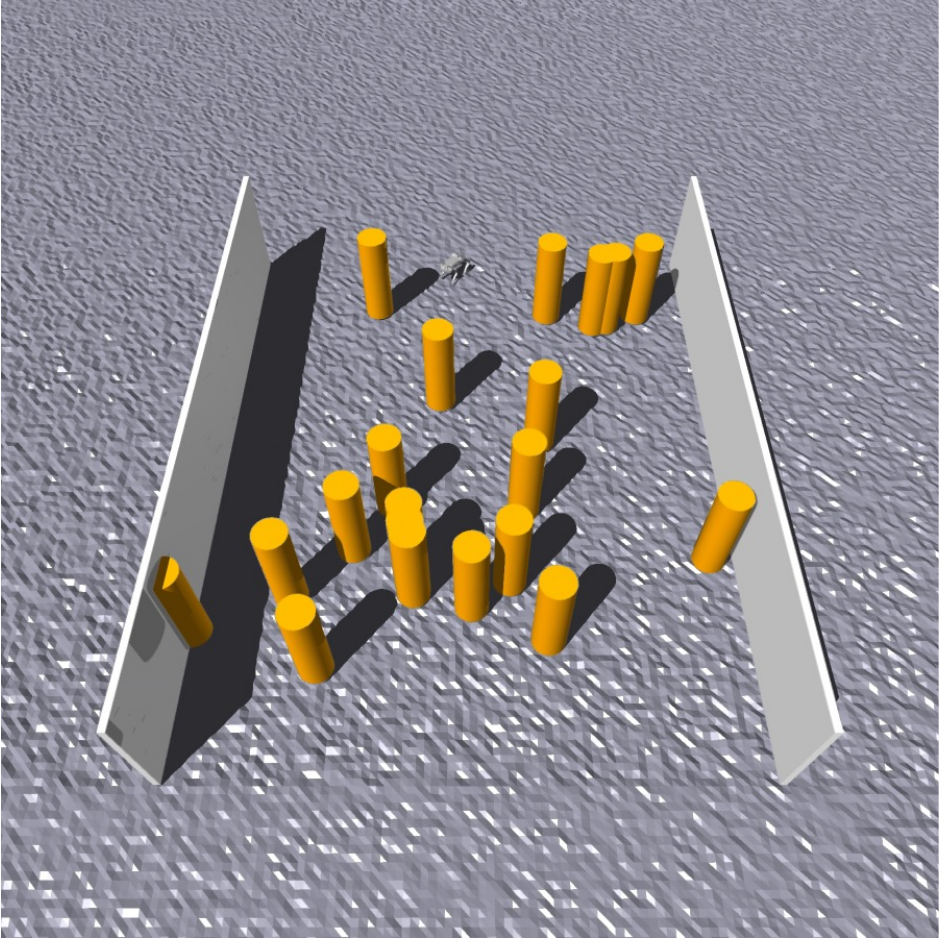}}
    \hfill
    \subfloat[Slow Environment]{\includegraphics[width=0.24\textwidth]{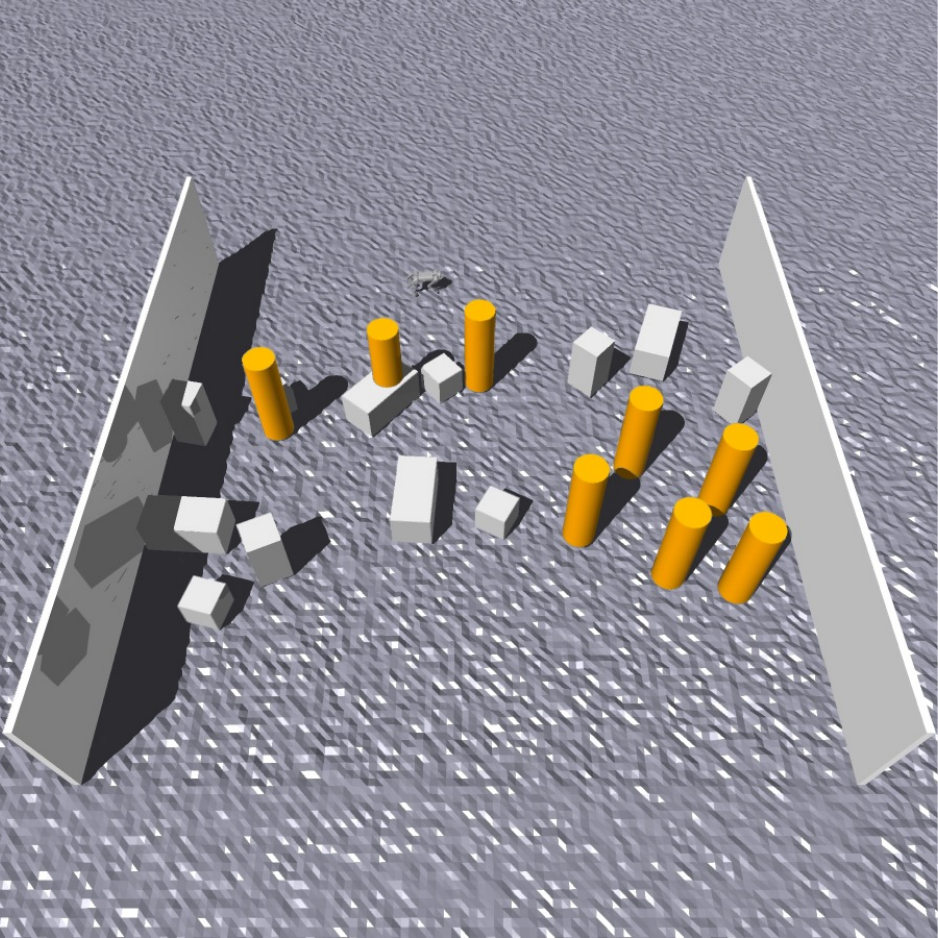}}
    \caption{Visualizations of the unseen test environments. Across all scenarios, yellow cylinders denote dynamic obstacles (simulated pedestrians). (a) Forest Environment: An unstructured outdoor setting where brown cylinders represent trees. (b) Office Environment: A structured indoor layout where gray cuboids and walls simulate furniture and corridors. (c) Square Environment: An open public plaza bounded by gray walls along two sides, characterized by high dynamic crowd density. {(d) Slow Environment: A low-speed dynamic interaction setting with gray box-shaped static obstacles, boundary walls, and reduced-speed dynamic obstacles.}}
    \label{fig:unseen_env}
\end{figure*}

We conducted extensive experiments in both simulated and real-world environments to address the following research questions:
\begin{enumerate}
    \item Can the VOP-Nav policy improve obstacle avoidance success rates in crowded dynamic environments while maintaining high locomotion speeds?
    \item Does the safe velocity region predicted by VOP-Net effectively contribute to collision avoidance?
    \item Can the VOP-Nav policy demonstrate robust sim-to-real transfer, achieving real-world performance comparable to simulation?
\end{enumerate}
{To answer these questions, we benchmark VOP-Nav against baselines covering five categories: the classical VO method ORCA~\cite{vandenberg2011}, the learning-based VO-aided method NavRL~\cite{xu2025}, the crowd navigation planner HEIGHT~\cite{liu2025a}, the quadruped crowd navigation method REASAN~\cite{yuan2025reasan}, and the agile quadruped navigation method ABS~\cite{he2024}.} Furthermore, we design a series of quantitative and qualitative analyses to verify the efficacy of the VOP-Net predictions. Finally, we evaluate the system's obstacle avoidance robustness and agile locomotion capabilities through rigorous real-world testing. The detailed experimental results are presented below.

\subsection{Evaluation in Unseen Simulated Environments}
To evaluate the generalization ability of VOP-Nav, we design four unseen test environments that differ substantially from the training setup. These environments emulate diverse real-world scenarios and introduce a clear domain shift in obstacle dynamics. In contrast to the linear motion model used during training, dynamic obstacles in the test environments are controlled by the Reciprocal Velocity Obstacle (RVO) algorithm~\cite{vandenberg2008}. RVO produces more realistic crowd behavior, where agents anticipate potential collisions and adjust their motion cooperatively.

We introduce a reactivity coefficient, $r_{\text{react}} \in [0,1]$, which controls how many dynamic agents account for the robot when computing reciprocal avoidance. Higher values of $r_{\text{react}}$ are used in denser environments to reflect the increased responsiveness typically observed in pedestrian crowds.

The absence of RVO dynamics during policy training is a deliberate design choice for three reasons. First, RVO incurs substantial computational overhead, which significantly reduces simulation throughput. Second, RVO agents often behave conservatively in crowded settings. When many agents interact, they may form local congestion and block navigable corridors. This freezing behavior restricts the robot’s exploration and leads to sparse rewards, slowing policy learning. Third, the inherent deceleration in RVO-based avoidance reduces the occurrence of high-speed interactions, limiting the robot’s exposure to aggressive and time-critical scenarios.

By training the policy with fast-moving linear obstacles and evaluating it against cooperative RVO-controlled crowds, we test robustness under previously unseen crowd dynamics while maintaining relevance to realistic crowd flows. The four test environments are illustrated in Fig.~\ref{fig:unseen_env} and described in detail below.

\begin{itemize}
    \item \textbf{Forest Environment:} Designed to simulate an unstructured outdoor setting, this environment contains 20 static obstacles and 6 dynamic obstacles ($r_\text{react} = 0.2$). The static obstacles are modeled as cylinders with randomized radii, distributed stochastically within grid partitions to approximate irregular tree placement. The workspace boundaries are open (wall-less). While the dynamic density is moderate, the combination of static occlusions and high-speed dynamic agents creates severe visibility constraints, challenging the policy's ability to react to sudden threats.

    \item \textbf{Office Environment:} This scenario approximates a structured indoor layout, featuring 30 static rectangular obstacles and 8 dynamic obstacles ($r_\text{react} = 0.4$) enclosed by walls. The rectangular obstacles are fixed in a predetermined arrangement to mimic furniture and corridors. The juxtaposition of dense static structures and dynamic agents with moderate reactivity creates a highly constrained space with narrow passages, rigorously testing the robot's precision maneuvering and collision avoidance in clutter.

    \item \textbf{Square Environment:} Simulating a crowded public plaza, this environment is devoid of static obstacles but populated with 20 dynamic agents ($r_\text{react} = 0.6$). Despite the high reactivity of the crowd, the sheer density of agents leads to emergent behaviors such as bottlenecking and trajectory intersection, which can dynamically block feasible paths. Although this represents an open space, boundary walls are retained to constrain the operational area, preventing the robot from exploiting trivial edge-following strategies.

    \item {\textbf{Slow Environment:} This scenario is designed to evaluate low-speed dynamic interactions. It contains 12 box-shaped static obstacles distributed randomly and 8 dynamic obstacles with reduced speeds ($0.6\,\text{m/s}$) and lower reactivity ($r_\text{react}=0.2$).}
\end{itemize}

Acknowledging that the complex dynamics in the unseen test environments necessitate more intricate evasion maneuvers and consequently longer traversal times, we extended the episode time limit from $10\,\text{s}$ to $15\,\text{s}$. Furthermore, the spatial dimensions of the test environments were expanded to accommodate the increased scene complexity. Unlike the training arena ($5 \times 10\,\text{m}^2$), the unseen environments feature larger footprints of either $8 \times 10\,\text{m}^2$ or $10 \times 10\,\text{m}^2$. A comprehensive comparison of the environmental parameters between the training and unseen test scenarios is presented in Table~\ref{tab:unseen_env}.

\begin{table*}[t]
    \centering
    \caption{Detailed Configurations of Training and Unseen Evaluation Environments}
    \label{tab:unseen_env}
    \begin{tabular}{ccccccccc}
        \toprule
        Name & Time & 
        \makecell[c]{Dynamic\\Obstacle} & 
        \makecell[c]{Static Cylinder\\Obstacle} & 
        \makecell[c]{Static Rectangular\\Obstacle} & 
        Wall & RVO use & $r_\text{react}$ & size($m^2$) \\
        \midrule
        Training Environment & 10 & 8 & 2 & 4 & True & False & 0 & $5 \times 10 $\\
        Forest Environment & 15 & 6 & 20 & 0 & False & True & 0.2 & $8 \times 10 $\\
        Office Environment & 15 & 8 & 0 & 30 & True & True & 0.4 & $10 \times 10 $\\
        Square Environment & 15 & 20 & 0 & 0 & True & True & 0.6 & $8 \times 10 $\\
        Slow Environment & 15 & 8 & 0 & 12 & True & True & 0.2 & $10 \times 10 $\\
        \bottomrule
    \end{tabular}
\end{table*}

\subsection{Baselines Evaluation}
\subsubsection{Baselines Selection and Adaptation}
\label{sec:baseline_selection}
{Identifying suitable baselines for direct comparison presents significant challenges due to fundamental architectural differences between existing approaches and our end-to-end quadruped framework. Most state-of-the-art crowded navigation methods and benchmarks are designed for wheeled robots with simple velocity-based action spaces rather than the high-dimensional joint control required by quadruped robots. We therefore adapt representative baselines from several categories to our simulation environments and make every effort to optimize their performance. Nevertheless, environment differences and, for high-level planners, compatibility with their locomotion models may still affect results. The selected baselines and their adaptations are detailed below.}

\begin{itemize}
    \item {\textbf{ORCA.} We implement an ORCA-based planner~\cite{vandenberg2011} with privileged simulator obstacle positions and velocities. For noncooperative obstacles, we set the robot's avoidance responsibility to 1.0, assigning full collision-avoidance responsibility to the robot. This baseline evaluates direct VO reasoning without perception errors.}

    \item {\textbf{NavRL.} NavRL~\cite{xu2025} combines a learning-based navigation policy with a VO safety shield that adjusts the policy's preferred command according to VO constraints. Both the policy and safety shield use privileged dynamic-obstacle states.}

    \item {\textbf{HEIGHT.} HEIGHT~\cite{liu2025a} is a learning-based interaction-aware crowd navigation method that uses agent-state inputs. We supply these inputs using privileged obstacle information. We use the authors' official pre-trained model and execute its high-level commands through the Rapid locomotion controller~\cite{margolis2024}.}

    \item {\textbf{REASAN.} REASAN~\cite{yuan2025reasan} is included as a modularized end-to-end reactive navigation method for quadruped robots in dynamic environments. It is closely related to our setting because it performs legged reactive navigation using onboard sensing only.}

    \item {\textbf{ABS.} ABS~\cite{he2024} is included as an agile end-to-end quadruped navigation baseline. {It combines an agile policy with a recovery policy selected by a reach-avoid value (RA-Value) network. Under normal conditions, the robot is controlled by the agile policy. When the RA-Value Network indicates potential collision risk, the recovery policy takes over and executes emergency avoidance or deceleration.} Since the RA-Value Network did not converge in our dynamic training environments, we reuse the authors' pre-trained RA-Value Network and recovery policy, and retrain only the agile policy in our environment to adapt it to dynamic obstacle interactions.}
\end{itemize}

{For the newly added ORCA, NavRL, and REASAN baselines, we use the same REASAN locomotion policy checkpoint~\cite{yuan2025reasan} to reduce differences in low-level execution models. Unless otherwise specified, all baselines use the standard episode duration. ORCA and NavRL are additionally evaluated with doubled episode duration (2T) to reduce timeout-driven failures. For HEIGHT, we report results with a $4\times$ extended episode duration (HEIGHT(4T)) because its high-level forward-velocity command is limited to $0.5\,\text{m/s}$, below that required in our navigation tests.}

\subsubsection{Results}
{We evaluate VOP-Nav and the baselines in Table~\ref{tab: baseline comparasion} across the training environment and four test environments, using success rate, collision rate, timeout rate, mean velocity ($\bar{v}$), and mean peak velocity ($\bar{v}_\text{peak}$) of successful trajectories.}

\begin{table*}[t]
    \centering
    \caption{Performance comparison of VOP-Nav against baselines across training and unseen environments.} 
    \label{tab: baseline comparasion}
    \begin{tabular}{ccccccc}
        \toprule
        Environment Name & Policy Name & Success Rate(\%)$\uparrow$& Collision Rate(\%)$\downarrow$& Timeout Rate(\%)$\downarrow$& $\bar{v}$ of Success(m/s)$\uparrow$& $\bar{v}_\text{peak}$ of Success(m/s)$\uparrow$\\
        \midrule
        \multirow{8}{*}{Training}   & {HEIGHT(4T)} & 0.17 & 29.05 & 70.79 & 0.48 & 0.65\\
                                    & ABS              & 53.16 & 43.57 & 3.26 & \textbf{2.29} & \textbf{3.42}\\
                                    & {ORCA}       & {35.09} & {\textbf{14.14}} & {50.77} & {1.50} & {2.71}\\
                                    & {ORCA(2T)}   & {59.89} & {22.73} & {17.38} & {1.21} & {2.65}\\
                                    & {NavRL}      & {25.71} & {15.01} & {59.28} & {1.37} & {2.52}\\
                                    & {NavRL(2T)}  & {71.84} & {22.55} & {5.61} & {1.13} & {2.49}\\
                                    & {REASAN}     & {43.49} & {48.10} & {8.41} & {1.61} & {2.44}\\
                                    & VOP-Nav          & \textbf{78.35} & 18.91 & \textbf{2.74} & 1.99 & 3.34\\
        \hline
        \multirow{8}{*}{Forest}     & {HEIGHT(4T)} & 26.65 & 20.00 & 53.35 & 0.46 & 0.66\\
                                    & ABS              & 76.07 & 23.42 & 0.50 & 2.47 & 3.44\\
                                    & {ORCA}       & {85.09} & {11.31} & {3.59} & {1.62} & {2.69}\\
                                    & {ORCA(2T)}   & {87.65} & {11.13} & {1.22} & {1.58} & {2.69}\\
                                    & {NavRL}      & {89.25} & {8.25} & {2.50} & {1.58} & {2.56}\\
                                    & {NavRL(2T)}  & {90.49} & {8.68} & {0.84} & {1.56} & {2.57}\\
                                    & {REASAN}     & {68.06} & {30.90} & {1.04} & {1.67} & {2.35}\\
                                    & VOP-Nav          & \textbf{94.48} & \textbf{5.42} & \textbf{0.10} & \textbf{2.55} & \textbf{3.56}\\
        \hline
        \multirow{8}{*}{Office}     & {HEIGHT(4T)} & 59.46 & 22.39 & 18.14 & 0.41 & 0.67\\
                                    & ABS              & 61.29 & 29.33 & 9.38 & \textbf{2.15} & 3.35\\
                                    & {ORCA}       & {47.64} & {\textbf{8.84}} & {43.52} & {1.33} & {2.72}\\
                                    & {ORCA(2T)}   & {62.01} & {13.08} & {24.90} & {1.05} & {2.69}\\
                                    & {NavRL}      & {65.43} & {11.45} & {23.12} & {1.23} & {2.52}\\
                                    & {NavRL(2T)}  & {81.07} & {13.60} & {5.34} & {1.13} & {2.52}\\
                                    & {REASAN}     & {64.07} & {33.09} & {\textbf{2.84}} & {1.59} & {2.46}\\
                                    & VOP-Nav          & \textbf{81.12} & 14.93 & 3.95 & 2.03 & \textbf{3.49}\\
        \hline
        \multirow{8}{*}{Square}     & {HEIGHT(4T)} & 3.75 & 23.69 & 72.56 & 0.44 & 0.67\\
                                    & ABS              & 53.11 & 44.19 & 2.70 & \textbf{2.24} & 3.30\\
                                    & {ORCA}       & {76.92} & {19.94} & {3.14} & {1.53} & {2.70}\\
                                    & {ORCA(2T)}   & {78.79} & {20.31} & {0.89} & {1.50} & {2.70}\\
                                    & {NavRL}      & {71.81} & {20.51} & {7.68} & {1.25} & {2.50}\\
                                    & {NavRL(2T)}  & {76.37} & {21.96} & {1.67} & {1.22} & {2.50}\\
                                    & {REASAN}     & {47.34} & {50.60} & {2.06} & {1.57} & {2.40}\\
                                    & VOP-Nav          & \textbf{83.67} & \textbf{15.67} & \textbf{0.65} & 1.96 & \textbf{3.35}\\
        \hline
        \multirow{8}{*}{Slow}       & {HEIGHT(4T)} & 47.22 & 29.64 & 23.14 & 0.41 & 0.68\\
                                    & ABS              & 83.92 & 13.78  & 2.30 & \textbf{2.30} & 3.34\\
                                    & {ORCA}       & {66.68} & {\textbf{2.26}} & {31.06} & {1.37} & {2.68}\\
                                    & {ORCA(2T)}   & {76.04} & {3.42} & {20.53} & {1.17} & {2.65}\\
                                    & {NavRL}      & {76.46} & {5.07} & {18.47} & {1.18} & {2.51}\\
                                    & {NavRL(2T)}  & {88.89} & {5.98} & {5.13} & {1.09} & {2.50}\\
                                    & {REASAN}     & {86.23} & {12.37} & {\textbf{1.40}} & {1.59} & {2.39}\\
                                    & VOP-Nav          & \textbf{92.24} & 6.08  & 1.69 & 2.19 & \textbf{3.42}\\
        \bottomrule
    \end{tabular}
\end{table*}
\begin{figure*}[t]
    \centering
    \subfloat[HEIGHT]{\includegraphics[width=0.16\textwidth]{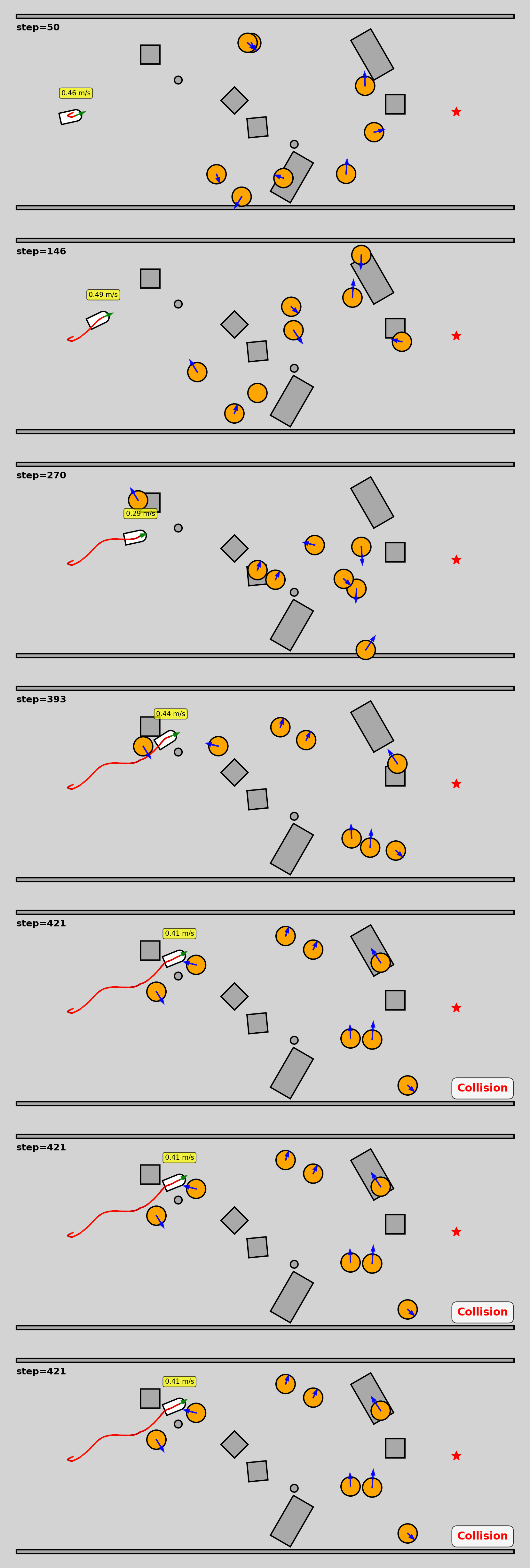}}
    \hfill
    \subfloat[ORCA]{\includegraphics[width=0.16\textwidth]{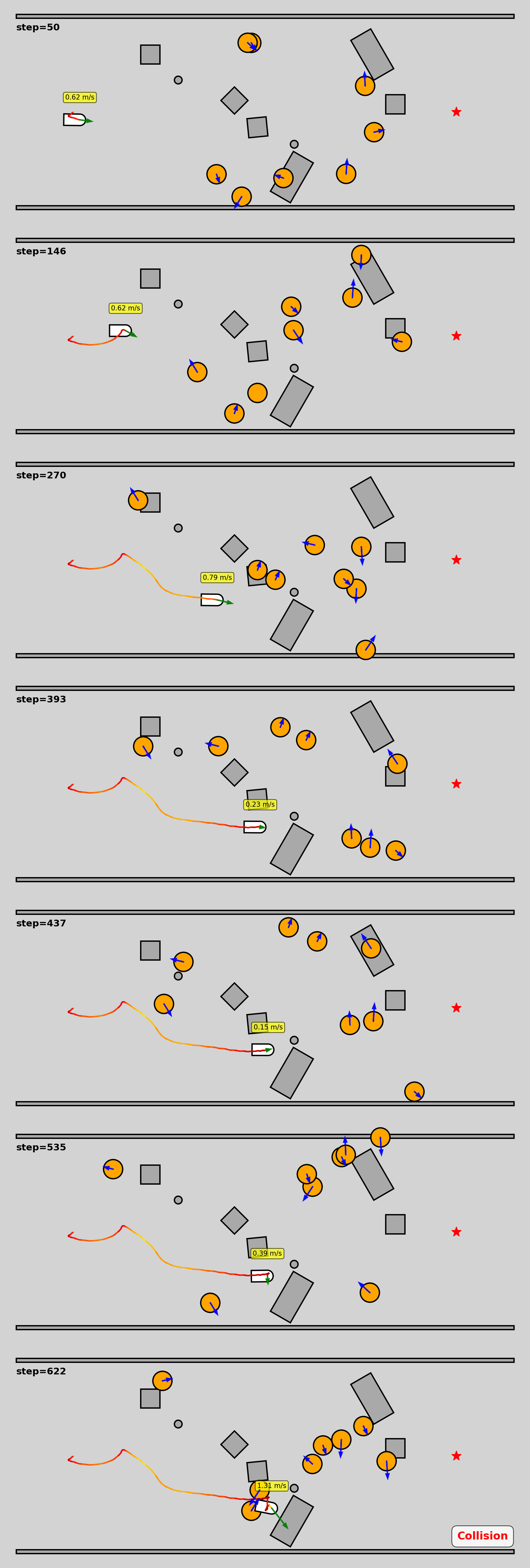}}
    \hfill
    \subfloat[NavRL]{\includegraphics[width=0.16\textwidth]{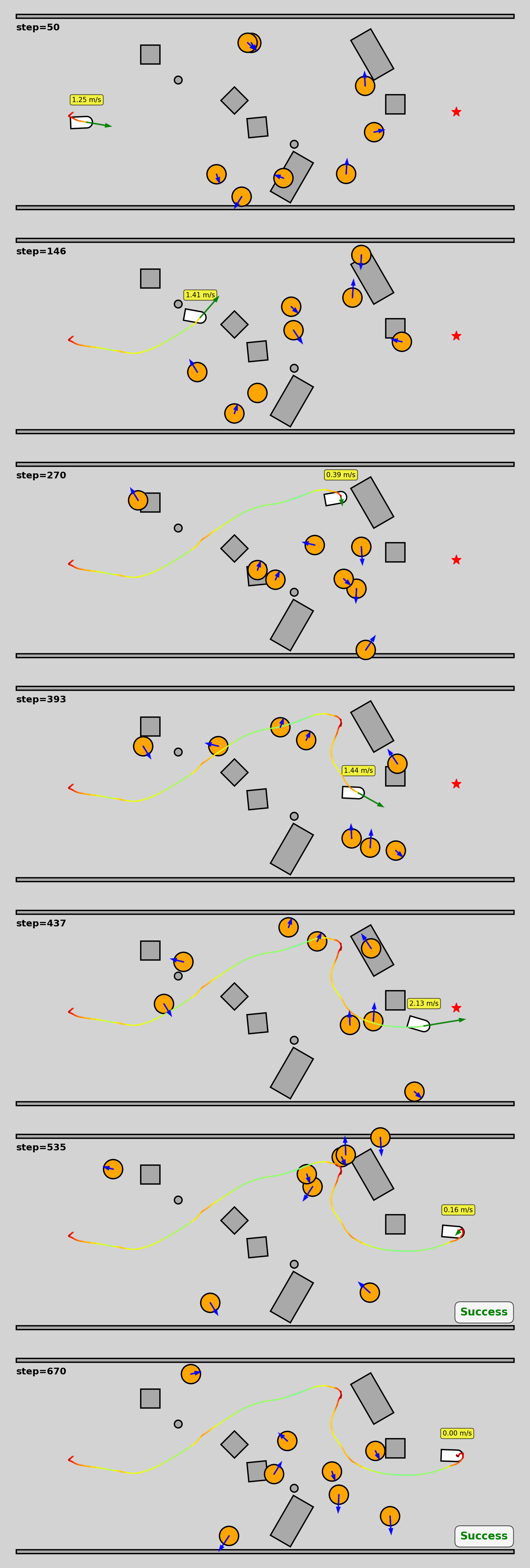}}
    \hfill
    \subfloat[REASAN]{\includegraphics[width=0.16\textwidth]{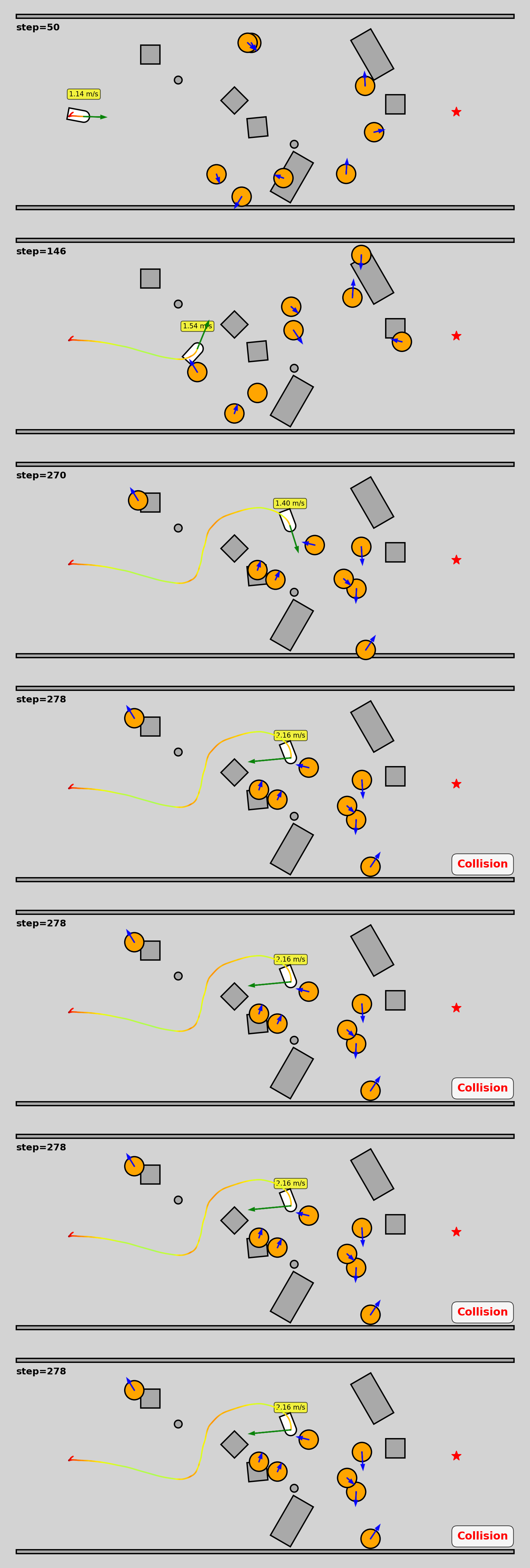}}
    \hfill
    \subfloat[ABS]{\includegraphics[width=0.16\textwidth]{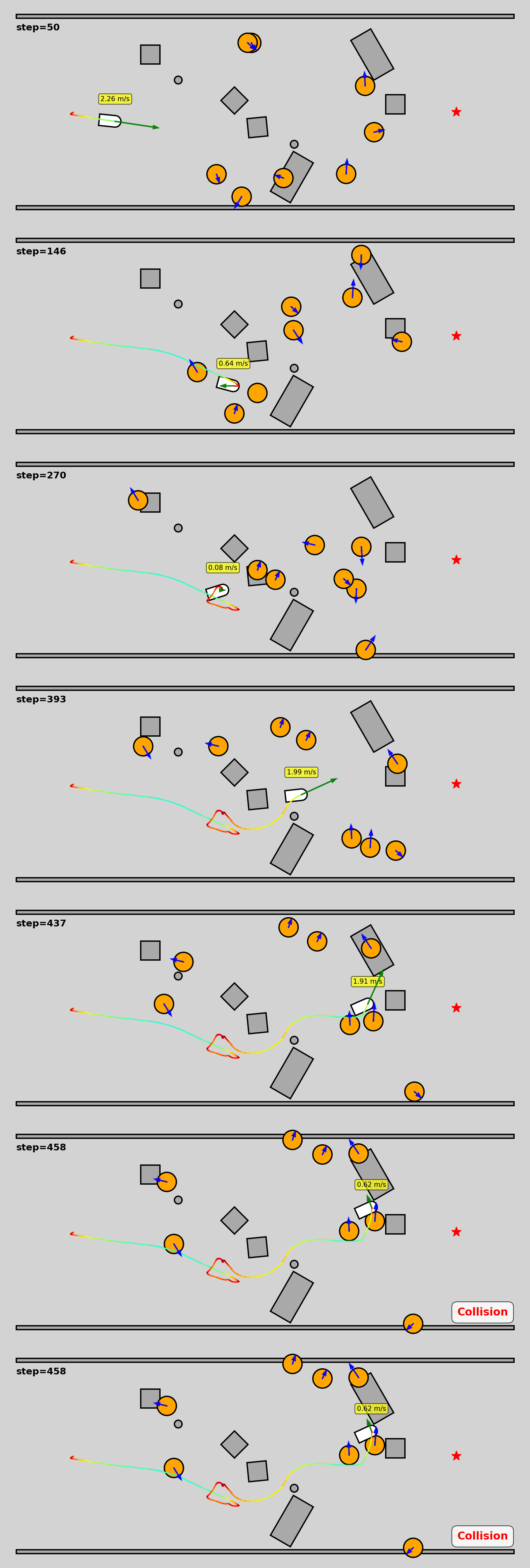}}
    \hfill
    \subfloat[VOP-Nav]{\includegraphics[width=0.16\textwidth]{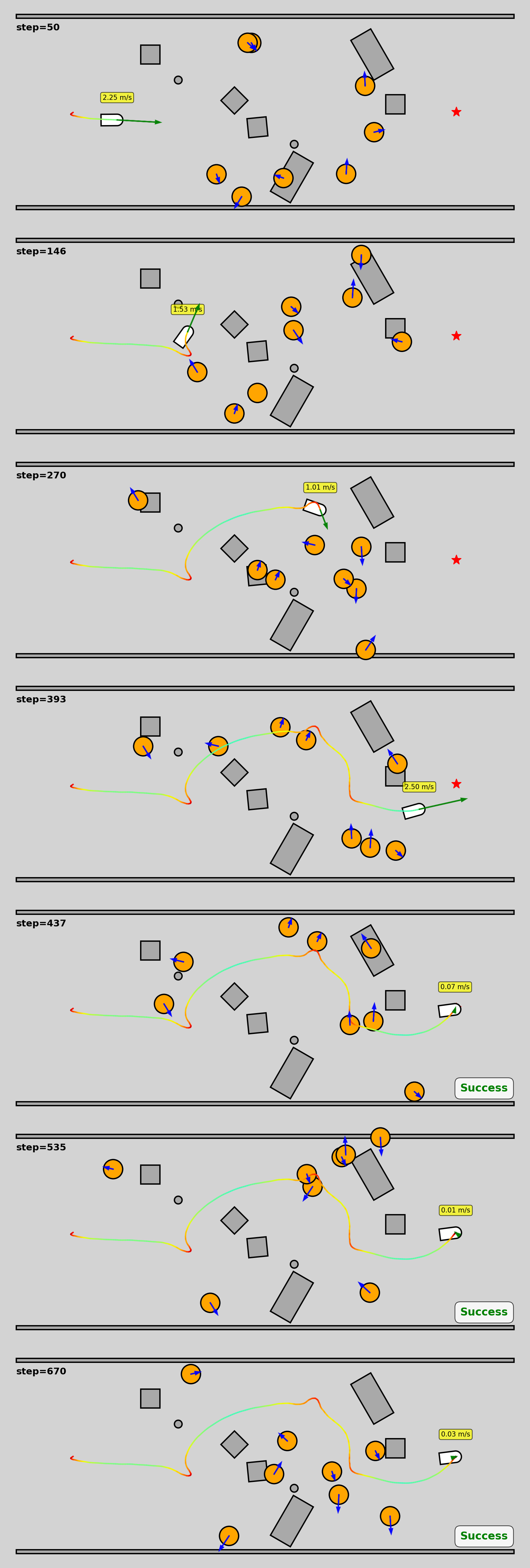}}
    \caption{Qualitative comparison of trajectories in the standardized test environment. The environment is modified from the training scenario with manually designated static obstacles (grey) to extend trajectory length, while dynamic obstacles (orange) are generated using identical random seeds. The robot is depicted as a white semi-circle (head) with a rectangular body. The yellow box above the robot displays the speed magnitude of the current frame. Arrows indicate velocity vectors for the robot (green) and dynamic obstacles (blue). The robot's trajectory is color-coded by speed, varying from red (low speed) to blue (high speed). The red star denotes the goal position.}
    \label{fig:Baseline}
\end{figure*}

As presented in Table~\ref{tab: baseline comparasion}, {VOP-Nav achieves the highest success rates across all five environments while maintaining competitive locomotion speeds.} The Training environment constitutes the most demanding scenario, characterized by restricted maneuvering space, high obstacle density, and aggressive dynamic obstacle velocities ($1.5\,\text{m/s}$). {In the Training environment, ORCA and NavRL achieve relatively low collision rates but exhibit high timeout rates. Doubling the episode duration improves their success rates, but neither surpasses VOP-Nav. REASAN and ABS exhibit higher collision rates, while HEIGHT records low success and high timeout rates even with an extended episode duration. In the Forest environment, ORCA and NavRL remain competitive, whereas REASAN achieves its strongest result in the lower-speed Slow environment. In the Office environment, NavRL(2T) nearly matches VOP-Nav in success rate when evaluated with twice the standard episode duration. VOP-Nav retains the highest success rate in all three environments.} The Square environment represents our primary evaluation focus, constituting a truly crowded dynamic scenario. {Here, HEIGHT records low success and high timeout rates, while ABS and REASAN exhibit high collision rates. ORCA and NavRL perform substantially better but still achieve lower success rates and higher collision rates than VOP-Nav.}

Analyzing the velocities of successful trajectories reveals distinct behavioral strategies. {HEIGHT, REASAN, ORCA, and NavRL use hierarchical architectures. Their measured speeds depend on both locomotion capabilities and planner output limits (peaks above these limits result from locomotion tracking errors). We therefore focus on the joint-level policies ABS and VOP-Nav. As shown in Table~\ref{tab: baseline comparasion}, VOP-Nav achieves peak speeds comparable to those of ABS, exceeding them in several environments, while its mean speed is generally lower.} These results indicate that VOP-Nav adapts its velocity to surrounding dynamic constraints, retaining high peak speeds when transient gaps are available while slowing down during hazardous interactions.

{Taken together, the results reveal a consistent pattern across paradigms. ORCA and NavRL achieve lower collision rates than ABS and REASAN across all environments, demonstrating the benefit of VO reasoning for dynamic avoidance. However, their lower success rates than VOP-Nav and high timeout rates in several environments also expose two main limitations of directly applying VO constraints to dense, noncooperative quadruped navigation:}

\begin{itemize}
    \item {\textbf{Conservative velocity selection.} Direct VO methods typically select the collision-free velocity closest to a goal-directed preferred velocity. This local objective does not account for longer-term detours, which can lead to hesitation, freezing, and timeouts.}
    \item {\textbf{Conditions for VO-based safety.} VO-based safety relies on accurate obstacle information and the availability of collision-free velocities. Even with full avoidance responsibility assigned to the robot, crowded scenes can substantially reduce the feasible velocity set, while non-cooperative obstacles may eliminate it entirely. Moreover, in a hierarchical architecture, velocity-tracking delay in the locomotion model can prevent a safe high-level command from taking effect in time, further reducing the safety of direct VO methods.}
\end{itemize}

{In real-world deployment, ORCA and NavRL must obtain obstacle information through detection and tracking modules, which introduce additional uncertainty. The HEIGHT results further show that conservative speed limits task completion in highly dense dynamic settings. VOP-Nav instead incorporates the safe velocity region into both policy training and inference, allowing its joint-level policy to combine VO-based guidance with learned navigation and execution dynamics.}

\subsubsection{Trajectory Analysis}
\label{sec:traj_analysis}
{Figure~\ref{fig:Baseline} illustrates representative trajectories of six methods in the standardized test environment. The HEIGHT policy moved slowly and collided at Step 421 because its conservative motion could not evade a fast-moving obstacle. {ORCA repeatedly slowed down, as discussed in the previous subsection, after entering the crowded region at Step 270. It then experienced prolonged freezing in a corner and collided at Step 622 when two noncooperative dynamic obstacles nearly eliminated the feasible velocity set.} NavRL moved more decisively under its learning-based high-level planner. After slowing sharply and adjusting its course at Step 270, it passed through a traversable gap and reached the goal at Step 535. This successful trajectory demonstrates the benefit of combining learning-based planning with a VO safety shield. However, its small clearance at Step 270 indicates that low-level locomotion tracking errors under loose coupling may reduce the safety margin of high-level commands. REASAN reacted promptly to a fast-moving obstacle but failed to decelerate in time when multiple obstacles appeared, colliding at Step 278. ABS's recovery policy enabled several emergency decelerations in the early phase, but the robot repeatedly entered crowded regions and was struck by a dynamic obstacle at Step 437 after failing to handle the dense interaction.}

{VOP-Nav instead decelerated and moved into open space at Steps 146 and 270 rather than entering crowded regions. It accelerated once the path became feasible and reached the goal at Step 437, earlier than NavRL, while the other four methods collided. This trajectory shows that VOP-Nav regulates both speed and direction to avoid hazardous interactions while exploiting feasible gaps.}

\subsubsection{{Training Stability Analysis}}
\label{sec:training_stability}
{To verify that the reported VOP-Nav performance is not an artifact of a single lucky training seed, we trained the VOP-Nav policy with eight different random seeds under the same Stage-3 settings. {Table~\ref{tab:training_stability} reports the mean and standard deviation across these seeds in all five environments. Across the Training, Forest, Square, and Slow environments, the multi-seed averages remain close to the corresponding main-checkpoint results in Table~\ref{tab: baseline comparasion}, with small standard deviations across seeds. For example, the success rate in the Training environment is 77.50\% ($\pm$2.05), compared with 78.35\% for the main checkpoint. Office is the main exception, with larger standard deviations in success rate (72.66\% $\pm$ 6.41) and collision rate (23.60\% $\pm$ 6.87). We attribute this reduced stability to the greater static occlusion in Office, which limits onboard observability and makes policy performance more sensitive to training initialization. The mean and peak velocities of successful trajectories are also consistent across seeds.}}

\begin{table*}[t]
    \centering
    \caption{{Multi-seed performance of VOP-Nav variants across training and unseen environments.}}
    \label{tab:training_stability}
    \setlength{\tabcolsep}{3pt}
    \begin{tabular}{ccccccc}
        \toprule
        {Environment Name} & {Policy Name} & {Success Rate(\%)$\uparrow$} & {Collision Rate(\%)$\downarrow$} & {Timeout Rate(\%)$\downarrow$} & {$\bar{v}$ of Success(m/s)$\uparrow$} & {$\bar{v}_\text{peak}$ of Success(m/s)$\uparrow$} \\
        \midrule
        \multirow{3}{*}{{Training}}  & {GT-VOP-Nav} & {\textbf{79.55 $\pm$ 2.22}} & {\textbf{14.20 $\pm$ 1.72}} & {6.24 $\pm$ 2.66} & {\textbf{1.95 $\pm$ 0.05}} & {\textbf{3.46 $\pm$ 0.06}} \\
                                    & {VOP-Nav} & {77.50 $\pm$ 2.05} & {19.59 $\pm$ 3.11} & {\textbf{2.91 $\pm$ 1.44}} & {1.93 $\pm$ 0.04} & {3.27 $\pm$ 0.03} \\
                                    & {50\%d-VOP-Nav} & {75.51 $\pm$ 3.04} & {21.41 $\pm$ 3.01} & {3.08 $\pm$ 1.33} & {1.94 $\pm$ 0.05} & {3.29 $\pm$ 0.09} \\
        \hline
        \multirow{3}{*}{{Forest}}     & {GT-VOP-Nav} & {95.18 $\pm$ 0.69} & {4.75 $\pm$ 0.67} & {\textbf{0.07 $\pm$ 0.03}} & {\textbf{2.79 $\pm$ 0.05}} & {\textbf{3.85 $\pm$ 0.07}} \\
                                    & {VOP-Nav} & {95.16 $\pm$ 0.52} & {4.71 $\pm$ 0.53} & {0.13 $\pm$ 0.05} & {2.60 $\pm$ 0.08} & {3.60 $\pm$ 0.07} \\
                                    & {50\%d-VOP-Nav} & {\textbf{95.30 $\pm$ 1.45}} & {\textbf{4.63 $\pm$ 1.44}} & {\textbf{0.07 $\pm$ 0.03}} & {2.65 $\pm$ 0.12} & {3.69 $\pm$ 0.13} \\
        \hline
        \multirow{3}{*}{{Office}}     & {GT-VOP-Nav} & {\textbf{79.65 $\pm$ 7.06}} & {\textbf{15.00 $\pm$ 3.81}} & {5.35 $\pm$ 4.30} & {\textbf{2.14 $\pm$ 0.12}} & {\textbf{3.61 $\pm$ 0.16}} \\
                                    & {VOP-Nav} & {72.66 $\pm$ 6.41} & {23.60 $\pm$ 6.87} & {\textbf{3.74 $\pm$ 1.15}} & {2.08 $\pm$ 0.18} & {3.47 $\pm$ 0.10} \\
                                    & {50\%d-VOP-Nav} & {71.40 $\pm$ 14.25} & {22.82 $\pm$ 13.76} & {5.78 $\pm$ 6.73} & {2.13 $\pm$ 0.14} & {3.53 $\pm$ 0.15} \\
        \hline
        \multirow{3}{*}{{Square}}     & {GT-VOP-Nav} & {79.12 $\pm$ 2.07} & {20.31 $\pm$ 2.04} & {0.57 $\pm$ 0.24} & {\textbf{2.08 $\pm$ 0.06}} & {\textbf{3.44 $\pm$ 0.09}} \\
                                    & {VOP-Nav} & {\textbf{82.71 $\pm$ 1.81}} & {\textbf{16.61 $\pm$ 1.92}} & {0.68 $\pm$ 0.22} & {1.97 $\pm$ 0.09} & {3.33 $\pm$ 0.05} \\
                                    & {50\%d-VOP-Nav} & {81.22 $\pm$ 1.94} & {18.23 $\pm$ 1.88} & {\textbf{0.54 $\pm$ 0.25}} & {2.00 $\pm$ 0.07} & {3.35 $\pm$ 0.09} \\
        \hline
        \multirow{3}{*}{{Slow}}       & {GT-VOP-Nav} & {\textbf{93.13 $\pm$ 1.24}} & {\textbf{4.50 $\pm$ 0.83}} & {2.37 $\pm$ 0.99} & {\textbf{2.16 $\pm$ 0.04}} & {\textbf{3.56 $\pm$ 0.07}} \\
                                    & {VOP-Nav} & {92.83 $\pm$ 1.91} & {6.21 $\pm$ 2.34} & {\textbf{0.97 $\pm$ 0.65}} & {2.14 $\pm$ 0.07} & {3.38 $\pm$ 0.05} \\
                                    & {50\%d-VOP-Nav} & {92.11 $\pm$ 2.19} & {6.70 $\pm$ 2.32} & {1.19 $\pm$ 0.59} & {\textbf{2.16 $\pm$ 0.06}} & {3.42 $\pm$ 0.09} \\
        \bottomrule
    \end{tabular}
\end{table*}

\subsection{Analysis of Velocity Obstacle Perception}
\label{sec:vop_analysis}
\subsubsection{{VOP-Net Evaluation}}
\begin{figure}[t]
    \centering
    \includegraphics[width=0.95\linewidth]{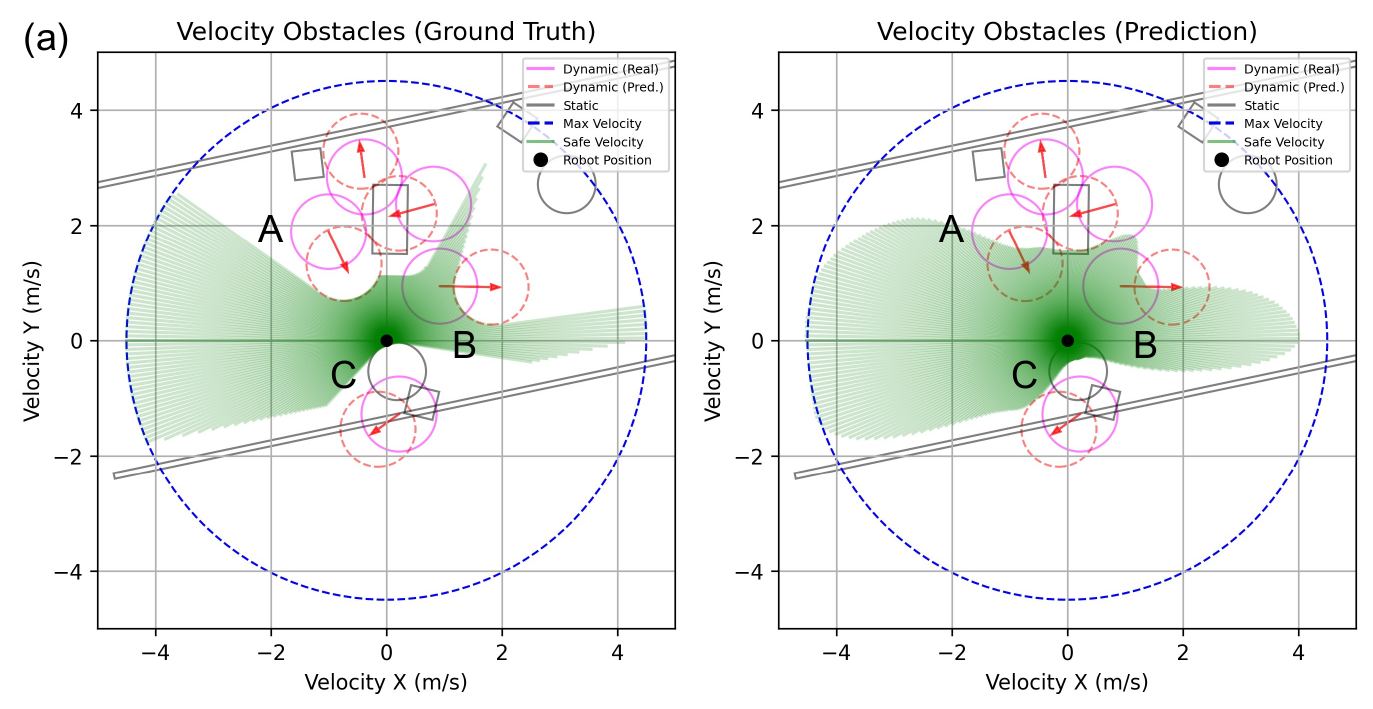}
    \vspace{0.2cm}
    \includegraphics[width=0.95\linewidth]{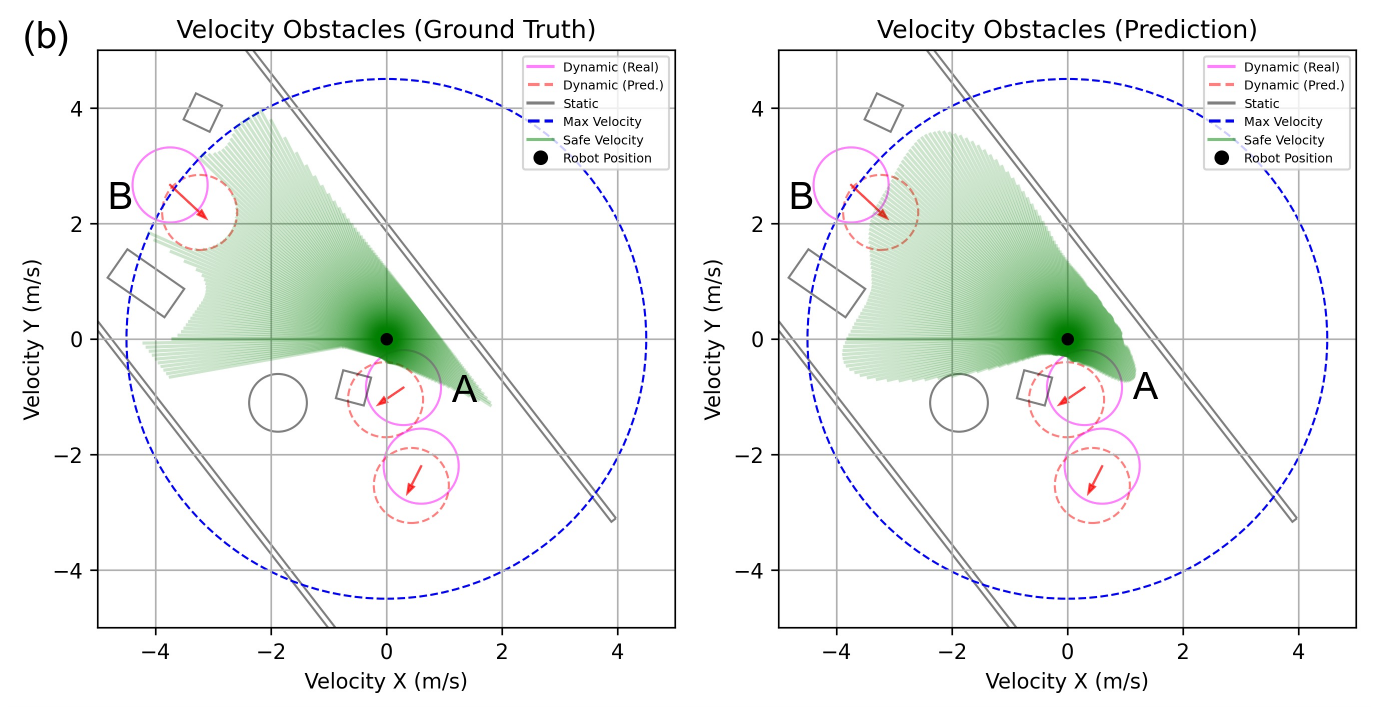}
    \caption{Visualization of safe velocity predictions in two scenarios (a) and (b). In each sub-figure, the left panel displays the ground truth, while the right panel shows the VOP-Net prediction. Magenta and red circles represent the current and next-step positions of dynamic obstacles, respectively, with red arrows indicating their velocity vectors. Static obstacles are depicted by black lines. Note that all cylindrical obstacles are visualized with their radii augmented by the robot's radius (configuration space). The blue outer ring marks the maximum speed limit ($v_{max}$), and the green regions represent the safe velocity region.}
    \label{fig:vop_analysis}
\end{figure}

{Our evaluation of VOP-Net covers quantitative prediction performance, the downstream impact of prediction errors, data efficiency, and qualitative prediction results.}

\noindent{\textbf{Quantitative Prediction Analysis.}} {VOP-Net achieves 0.808 IoU and 0.172 MSE on the validation sets. We further evaluate online prediction quality during final policy rollouts in the Training environment and four unseen environments. At each step, the ground-truth safe velocity region is computed from privileged simulator obstacle states using Algorithm~\ref{alg:1} and compared with the VOP-Net prediction from onboard observations. We use states between 5\% and 95\% of the robot's start-to-goal progress along the x-axis in successful trajectories. This sampling focuses on interaction-heavy segments, so online prediction quality is expected to be lower than the validation metrics. In addition to IoU and MSE, we use false-safe and false-unsafe rates to distinguish over-optimistic and over-conservative errors. Let $\ell^{\mathrm{pred}}_j$, $\ell^{\mathrm{gt}}_j$, and $\ell^{\cap}_j$ denote the lengths of the predicted safe interval, the ground-truth safe interval, and their intersection along direction $j$, respectively. $\mathrm{FS}$ measures the fraction of the predicted safe interval length unsupported by the ground truth, whereas $\mathrm{FU}$ measures the fraction of the ground-truth safe interval length omitted by the prediction. The two rates are computed as}
\begin{equation}
{
\mathrm{FS} =
\frac{\sum_j \left(\ell^{\mathrm{pred}}_j - \ell^{\cap}_j\right)}
     {\sum_j \ell^{\mathrm{pred}}_j},
\quad
\mathrm{FU} =
\frac{\sum_j \left(\ell^{\mathrm{gt}}_j - \ell^{\cap}_j\right)}
     {\sum_j \ell^{\mathrm{gt}}_j}.
}
\end{equation}
{Table~\ref{tab:vopnet_online} shows that VOP-Net maintains reasonable prediction quality across environments. Square yields the lowest IoU and the highest FS, making the dense Square environment the most challenging evaluated setting for safe velocity region prediction. More generally, FS tends to dominate in dense settings, whereas FU is more prominent in less crowded settings. This pattern shows that predictions during dense interactions are more prone to safety-critical over-optimistic errors, while errors in less crowded settings tend to produce more conservative guidance.}

\begin{table}[t]
    \centering
    \caption{{Online VOP-Net prediction metrics.}}
    \label{tab:vopnet_online}
    \resizebox{0.95\linewidth}{!}{\begin{tabular}{cccccc}
        \toprule
        {Environment Name} & {Model Name} & {IoU$\uparrow$} & {MSE$\downarrow$} & {FS(\%)$\downarrow$} & {FU(\%)$\downarrow$} \\
        \midrule
        \multirow{2}{*}{{Training}} & {VOP-Net}       & {\textbf{0.7005}} & {\textbf{0.2972}} & {16.49} & {\textbf{10.21}} \\
                                        & {50\%d-VOP-Net} & {0.6958} & {0.2986} & {\textbf{15.71}} & {11.80} \\
        \hline
        \multirow{2}{*}{{Forest}}   & {VOP-Net}       & {\textbf{0.6883}} & {\textbf{0.5581}} & {13.28} & {\textbf{16.22}} \\
                                        & {50\%d-VOP-Net} & {0.6859} & {0.5621} & {\textbf{11.66}} & {18.41} \\
        \hline
        \multirow{2}{*}{{Office}}   & {VOP-Net}       & {\textbf{0.7792}} & {\textbf{0.1667}} & {8.52}  & {\textbf{8.61}} \\
                                        & {50\%d-VOP-Net} & {0.7757} & {0.1684} & {\textbf{8.34}}  & {9.40} \\
        \hline
        \multirow{2}{*}{{Square}}   & {VOP-Net}       & {\textbf{0.5520}} & {\textbf{0.5368}} & {23.92} & {\textbf{14.44}} \\
                                        & {50\%d-VOP-Net} & {0.5429} & {0.5371} & {\textbf{23.07}} & {16.52} \\
        \hline
        \multirow{2}{*}{{Slow}}     & {VOP-Net}       & {\textbf{0.7891}} & {\textbf{0.2309}} & {8.07}  & {\textbf{10.12}} \\
                                        & {50\%d-VOP-Net} & {0.7866} & {0.2320} & {\textbf{7.47}}  & {11.19} \\
        \bottomrule
    \end{tabular}}
\end{table}

\noindent{\textbf{Downstream Impact of Prediction Errors.}} {To assess the downstream impact of VOP-Net prediction errors, we train GT-VOP-Nav with eight random seeds under the same Stage-3 settings. GT-VOP-Nav replaces VOP-Net predictions with ground-truth safe velocity regions for both policy input and reward supervision. As shown in Table~\ref{tab:training_stability}, GT-VOP-Nav improves success rates and substantially reduces collision rates in most environments, while achieving higher average and peak velocities in all five. This pattern indicates that prediction errors primarily affect collision avoidance rather than task completion. The largest gap occurs in Office, where static occlusion challenges onboard prediction. Square remains the main exception in success and collision performance, possibly because rapidly changing frame-wise ground-truth regions provide less smooth guidance under extreme crowding. Overall, VOP-Net uses only onboard observations yet preserves most of the navigation performance achieved with ground-truth safe velocity information. Its remaining errors mainly limit collision avoidance in perception-challenging scenes.}

\noindent{\textbf{Data-Scale Analysis.}} {We next examine how the VOP-Net training data scale affects prediction quality and downstream navigation. 50\%d-VOP-Net is trained on a random half of the 10,000 data files, and the corresponding 50\%d-VOP-Nav policy is trained with eight random seeds under the same Stage-3 settings. On the validation sets, 50\%d-VOP-Net achieves 0.806 IoU and 0.174 MSE, nearly matching the full-data model. Table~\ref{tab:vopnet_online} shows similarly small online changes in IoU and MSE but consistently lower FS and higher FU, indicating a more conservative estimate that omits more feasible velocities. At the policy level, Table~\ref{tab:training_stability} shows similar mean performance but higher success-rate standard deviations in all five environments, with Office most affected. Overall, halving the VOP-Net training data does not substantially reduce prediction accuracy or average downstream navigation success rates, but it reduces policy stability across random seeds.}

\noindent{\textbf{Qualitative Analysis.}} Specific case studies further demonstrate the network's capability to infer dynamic motion, as illustrated in Fig.~\ref{fig:vop_analysis}. In Case (a), the robot faces multiple dynamic obstacles (labeled A and B) on its left and is positioned extremely close to a static cylindrical obstacle (C) on its right. VOP-Net accurately captures the VO boundary of obstacle C; although there is a slight intrusion of predicted safe velocities into the collision zone, the RL policy will naturally avoid these marginal high-risk areas. Regarding the dynamic obstacles, the network exhibits distinct behaviors based on motion context. For the approaching obstacle A, while the prediction is not perfectly precise—with safe velocities overlapping a large portion of A's future collision region—the partial exclusion demonstrates that the network has successfully recognized the approaching threat. Conversely, for obstacle B, VOP-Net correctly identifies that it is moving away, allowing the collision region corresponding to the current frame to be fully covered by the predicted safe velocities.

A similar pattern is observed in Case (b). VOP-Net successfully identifies the motion status of the nearest dynamic object A and demonstrates awareness of the distant approaching object B. Furthermore, the prediction for the two nearby static obstacles is flawless without any erroneous overlap. These cases confirm that despite minor imperfections in extreme proximity, the safe velocities predicted by VOP-Net possess high reference value, effectively distinguishing between approaching threats and receding targets to support safe navigation.

\subsubsection{Visualization of Velocity Obstacle Perception}
\begin{figure*}[t]
    \centering
    \includegraphics[width=1\textwidth]{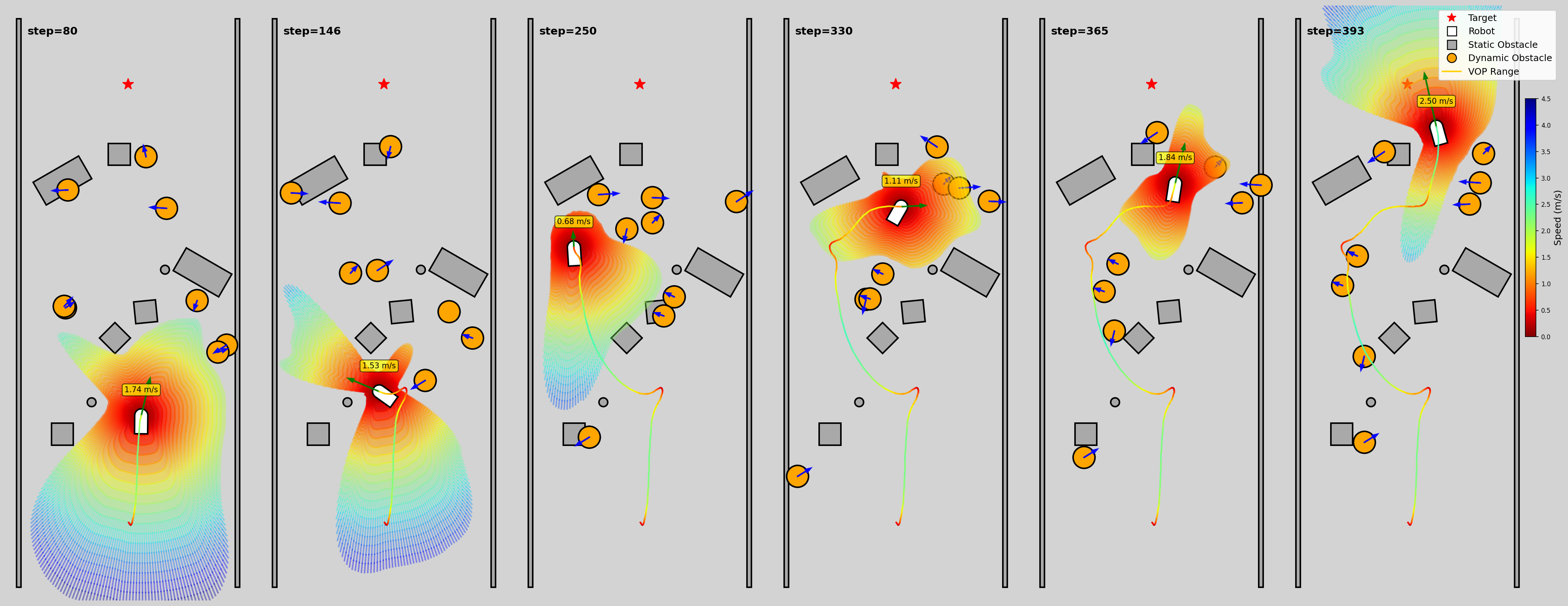}
    \caption{Visualization of velocity obstacle perception outputs along the VOP-Nav trajectory analyzed in Sec.~\ref{sec:traj_analysis}. The colored regions represent the predicted feasible velocity space ($\mathbf{V}_{\text{pred}}$) at key steps, where the color gradient indicates the magnitude of the safe speed. This visualization illustrates the dynamic evolution of the perception results during the navigation process.}
    \label{fig:horizon_voprl}
\end{figure*}
To intuitively demonstrate how the perception module guides navigation in complex scenarios, we visualized the evolution of the safe velocity region along the same trajectory as the VOP-Nav trial shown in Fig.~\ref{fig:Baseline}. As illustrated in Fig.~\ref{fig:horizon_voprl}, the predicted safe velocities are rendered using a color gradient consistent with the speed magnitude.

The visualization reveals the agent's decision-making logic at critical moments. At step $80$, the robot moves in an open area, where the safe velocity region is wide. At step $146$, approaching dynamic obstacles significantly reduce the feasible velocity space. At this point, the VOP module detects a clear split in the velocity space: a right-side path with lower safe speeds and a left-side path that supports higher velocities. Based on this information, the robot chooses the left direction, successfully avoiding the hazardous region in which the other baselines become trapped (see Sec.~\ref{sec:traj_analysis}). During the dense interaction phase (steps $250-330$), the robot behaves conservatively by slowing down and following surrounding agents. Notably, at step $330$, the predicted safe velocity region overlaps with the current position of a dynamic obstacle. This indicates that VOP-Net correctly identifies the obstacle as receding (moving away), thereby allowing the robot to utilize that space safely. Finally, as the robot approaches the exit at step $365$, the module detects a viable gap while maintaining awareness of two approaching threats on the right. By step $393$, the robot successfully clears the crowd, indicated by the expansive safe velocity field in the visualization.

\subsubsection{Ablation Study}
\begin{table*}[t]
    \centering
    \caption{Ablation study of different ways of incorporating VOP-Net predictions into the navigation policy.}
    \label{tab: ablation study}
    \begin{tabular}{ccccccc}
        \toprule
        Environment Name & Policy Name & Success Rate(\%)$\uparrow$& Collision Rate (\%)$\downarrow$& Timeout Rate(\%)$\downarrow$& $\bar{v}$ of Success(m/s)$\uparrow$& $\bar{v}_\text{peak}$ of Success(m/s)$\uparrow$\\
        \midrule
        \multirow{4}{*}{Training}  & {Basic} & {63.76} & {35.31} & {0.93} & {2.14} & {3.24}\\
                                   & Only reward & 61.14 & 27.22 & 11.64 & 1.85 & 3.23\\
                                   & Only input  & 68.77 & 30.56 & \textbf{0.67} & \textbf{2.39} & \textbf{3.78}\\
                                   & \textbf{VOP-Nav} & \textbf{78.35} & \textbf{18.91} & 2.74 & 1.99 & 3.34\\
        \hline
        \multirow{4}{*}{Forest} & {Basic} & {83.55} & {16.34} & {0.11} & {2.29} & {3.26}\\
                                & Only reward & \textbf{94.55} & \textbf{5.25} & 0.20 & 2.32 & 3.43\\
                                & Only input  & 84.07 & 15.61 & 0.32 & \textbf{2.99} & \textbf{4.14}\\
                                & \textbf{VOP-Nav} & 94.48 & 5.42 & \textbf{0.10} & 2.55 & 3.56\\
        \hline
        \multirow{4}{*}{Office} & {Basic} & {56.65} & {42.75} & {\textbf{0.60}} & {1.97} & {3.09}\\
                                & Only reward & 36.62 & 16.66 & 46.73 & 1.79 & 3.47\\
                                & Only input  & 72.17 & 27.24 & \textbf{0.60} & \textbf{2.37} & \textbf{3.80}\\
                                & \textbf{VOP-Nav} & \textbf{81.12} & \textbf{14.93} & 3.95 & 2.03 & 3.49\\
        \hline
        \multirow{4}{*}{Square} & {Basic} & {66.66} & {32.79} & {0.55} & {2.01} & {3.10}\\
                                & Only reward & 65.42 & 30.18 & 4.40 & 1.80 & 3.27\\
                                & Only input  & 71.52 & 28.12 & \textbf{0.36} & \textbf{2.41} & \textbf{3.69}\\
                                & \textbf{VOP-Nav} & \textbf{83.67} & \textbf{15.67} & 0.65 & 1.96 & 3.35\\
        \hline
        \multirow{4}{*}{Slow}   & {Basic} & {89.72} & {9.92}  & {0.36} & {2.06} & {3.10}\\
                                & Only reward & 86.66 & 6.58 & 6.76 & 1.92 & 3.23\\
                                & Only input  & 90.12 & 9.61 & \textbf{0.27} & \textbf{2.50} & \textbf{3.80}\\
                                & \textbf{VOP-Nav} & \textbf{92.24} & \textbf{6.08}  & 1.69 & 2.19 & 3.42\\
        \bottomrule
    \end{tabular}
\end{table*}

{Table~\ref{tab: ablation study} compares the Basic policy without any VOP component, the ``Only Reward'' and ``Only Input'' configurations, and the full VOP-Nav policy. Relative to Basic, VOP-Nav achieves higher success rates and lower collision rates in all five environments, demonstrating the overall benefit of the VOP-Nav integration. It also achieves the highest success rate and lowest collision rate in four environments, while ``Only Reward'' performs marginally better in Forest.}

{The ``Only Reward'' configuration achieves lower success rates than Basic in four environments and generally exhibits lower average speeds and higher timeout rates. Without safe velocity region input, the reward alone tends to produce conservative behavior that limits progress in complex scenes. Forest is the main exception, where ``Only Reward'' performs comparably to VOP-Nav, consistent with conservative behavior being sufficient in a less constrained environment. The ``Only Input'' configuration outperforms ``Only Reward'' in success rate in four environments and outperforms Basic in all five. It also achieves the highest average and peak velocities across all environments. However, its collision rates remain higher than those of VOP-Nav in every environment, showing that safe velocity region input provides the capability for agile motion but does not provide comparable safety without reward regulation.}

{By using VOP-Net predictions as input during inference and as a reward signal during training, VOP-Nav reduces speed relative to ``Only Input'' while increasing success rates and reducing collision rates across all environments. These results show that the input provides safe velocity information, while the reward regulates its use, yielding the strongest overall balance between progress and safety.}

\subsection{Real-world Experiments}
\subsubsection{Hardware setup} 
Our experimental platform is built upon the Unitree Go2 quadruped robot. The perception suite comprises a Livox Mid-360 LiDAR and an Intel RealSense D435i depth camera. For data processing, we extract 2D laser scans from the LiDAR at $60\,\text{Hz}$ and depth images from D435i at $60\,\text{Hz}$, with the latter downsampled to a resolution of $22 \times 40$ to match the simulation configuration.

Although VOP-Nav {does not require map-based planning}, reliable {localization} is required to obtain the robot's linear velocity and relative goal position. For indoor experiments, we utilize a motion capture system for precise robot localization. For outdoor scenarios, we employ Fast-LIO2~\cite{xu2022}, a robust LiDAR-inertial odometry (LIO) algorithm. While simple local odometry could theoretically suffice, we adopted the LIO-based approach due to its ease of implementation given our existing sensor payload, allowing for robust {localization} without necessitating additional hardware.

The computational workload is distributed across two onboard modules: an NVIDIA Jetson Orin Nano (8GB) hosts the VOP-Net and sensor drivers, while an Intel N100 mini-PC executes the VOP-Nav policy. Both VOP-Net predictions and policy inference operate synchronously at $50\,\text{Hz}$, ensuring responsive high-level control, while the underlying PD controller operates at $200\,\text{Hz}$.

\subsubsection{Experiments}
\begin{figure}[htbp]
  \centering
  \includegraphics[width=0.95\columnwidth]{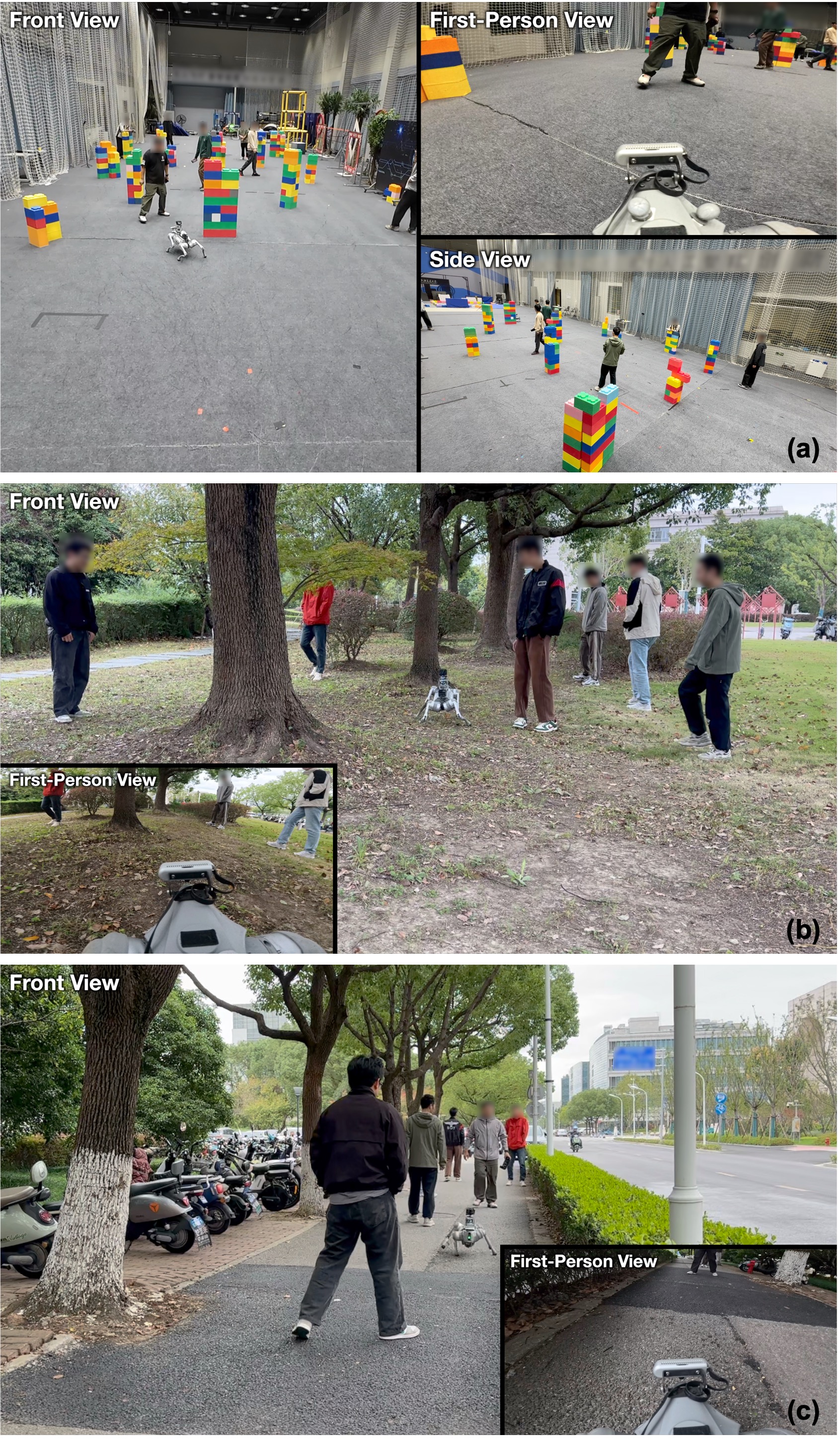}
  
  \caption{Experimental environments used in our evaluation. (a) The indoor scenario with static and dynamic obstacles. (b) The outdoor grove scenario with uneven terrain.  (c) The outdoor narrow path scenario.}
  \label{fig:realworld_environments}
\end{figure}
Due to the aggressiveness of our policy, despite its inherent safety, we conducted experiments in controlled settings to prevent public alarm. We carried out experiments in both indoor and outdoor environments, as shown in Fig. \ref{fig:realworld_environments}. In the indoor environment (Fig.~\ref{fig:realworld_environments}(a)), the test area measured approximately 20 m in length and 8 m in width. We constructed 12 static obstacles using colored bricks. Seven participants moved freely within the test area and were instructed to intentionally obstruct the robot by moving into its intended path. For outdoor testing, we selected a narrow path and a small grove, with eight participants moving between the start and end points to serve as dynamic obstacles. The grove environment (Fig.~\ref{fig:realworld_environments}(b)) was generally open but contained static obstacles like trees of various sizes and bushes, along with significant terrain variations, including elevations and depressions. Participants moved freely within this area and were permitted to actively interfere with the robot. The narrow path environment (Fig.~\ref{fig:realworld_environments}(c)) had a traversable width of approximately 3 m and a length of 20 m. This environment contained static obstacles such as trees, shrubbery, and non-motorized vehicles, and featured slightly uneven terrain. During these trials, all participants moved in scattered straight lines facing the robot and were permitted to actively interfere with it to simulate daily sidewalk scenarios. 

The experimental results are presented in Table~\ref{tab:experiment_results}. We conducted 15 trials in the indoor environment and achieved a $100\%$ success rate, demonstrating the stability of VOP-Nav in controlled real-world deployment. We also performed 15 trials in outdoor environments, where the robot completed 12 trials successfully, with two failures and one timeout. {The two outdoor failures were mainly associated with localization uncertainty}. The timeout occurred in an extremely narrow passage, where the robot prioritized safety by waiting for a feasible path rather than executing a risky maneuver. These results show that the VOP-Nav policy can be deployed zero-shot on the physical quadruped and operate safely in both indoor and outdoor dynamic scenarios.
\begin{table}[htbp]
    \centering
    \caption{Quantitative Results of Real-World Experiments}
    \label{tab:experiment_results}
    \setlength{\tabcolsep}{3pt}
    \begin{tabular}{lcccccc}
        \toprule
        Scenario & Trials & Success & Collision & Timeout & $\bar{v}_\text{peak}$(m/s) & $\bar{v}$(m/s) \\
        \midrule
        Indoor & 15 & \textbf{15} & \textbf{0} & \textbf{0} & 2.462 & \textbf{1.422} \\
        Outdoor & 15 & 12 & 2 & 1 & \textbf{2.493} & 1.370 \\
        \bottomrule
    \end{tabular}
\end{table}

\subsection{{Sim-to-Real Analysis}}
{Compared with its performance in simulation, VOP-Nav exhibits lower average traversal speeds in real-world deployment, as reported in Table~\ref{tab:experiment_results}. This gap is expected during sim-to-real transfer and does not undermine the real-world feasibility of VOP-Nav. The robot completed all indoor trials without collision and reached the goal in most outdoor trials. We attribute the gap mainly to the following factors.}

\noindent{\textbf{Scene and agent behavior mismatch.}}
{The simulated dynamic obstacles are simplified geometric agents with prescribed linear, stochastic, or RVO-style motion patterns, whereas real trials involve human participants who may stop, turn, accelerate, or deliberately obstruct the robot. These more diverse behaviors create more complex local interactions, often requiring the robot to decelerate or wait more frequently than in simulation.}

\noindent{\textbf{Onboard sensing mismatch.}}
{The Livox Mid-360 used in our system is a non-repetitive scanning 3D LiDAR. To obtain the 40-dimensional planar LiDAR input required by the policy, we use the LiDAR IMU to compensate for robot attitude, retain points near a plane parallel to the ground, and bin the remaining points by angle. The LiDAR operates at $60~\text{Hz}$, which reduces the average number of points per frame. Together with planar filtering, this can leave some angular sectors without valid points. We fill the corresponding distance measurements with the maximum range. The D435i depth images also undergo hole filling and downsampling before being fed to the policy. These sensing and preprocessing differences, together with the irregular geometry and surface properties of real obstacles, introduce observation shifts relative to simulation and may occasionally affect the policy's actions.}

\noindent{\textbf{Localization uncertainty.}}
{VOP-Nav relies on the input of relative goal position and robot linear velocity. Indoor trials use a motion-capture system and therefore provide accurate localization, which is consistent with the stable indoor performance. Outdoor trials rely on Fast-LIO2, and the aggressive motion of the robot can lead to accumulated odometry drift. When the drift becomes severe, the relative-goal observation and velocity estimate may become inaccurate, which can lead to failure. The two outdoor failures observed in our experiments were mainly associated with this localization uncertainty.}

\noindent{\textbf{Execution mismatch.}}
{Real deployment also introduces differences in locomotion execution, including computation and control latency, actuator response, foot wear, ground friction, and uneven terrain. These factors reduce the achievable motion performance compared with the simulator. Nevertheless, the high-speed test in Sec.~\ref{sec:extreme_robustness} shows that the real robot can still reach a peak speed above $3~\text{m/s}$ in open space, indicating that the lower speeds in crowded trials mainly reflect interaction and deployment constraints rather than a loss of basic locomotion capability.}

{Overall, the sim-to-real gap is mainly caused by scene-behavior mismatch, onboard sensing compromises, {localization uncertainty}, and execution mismatch. These factors explain the lower real-world speed and the few outdoor failures, while the indoor and outdoor results still demonstrate practical zero-shot deployment of VOP-Nav on the physical quadruped.}

\subsection{Extreme Performance and Robustness Analysis}
\label{sec:extreme_robustness}

\begin{figure}[t]
    \centering
    \includegraphics[width=0.95\columnwidth]{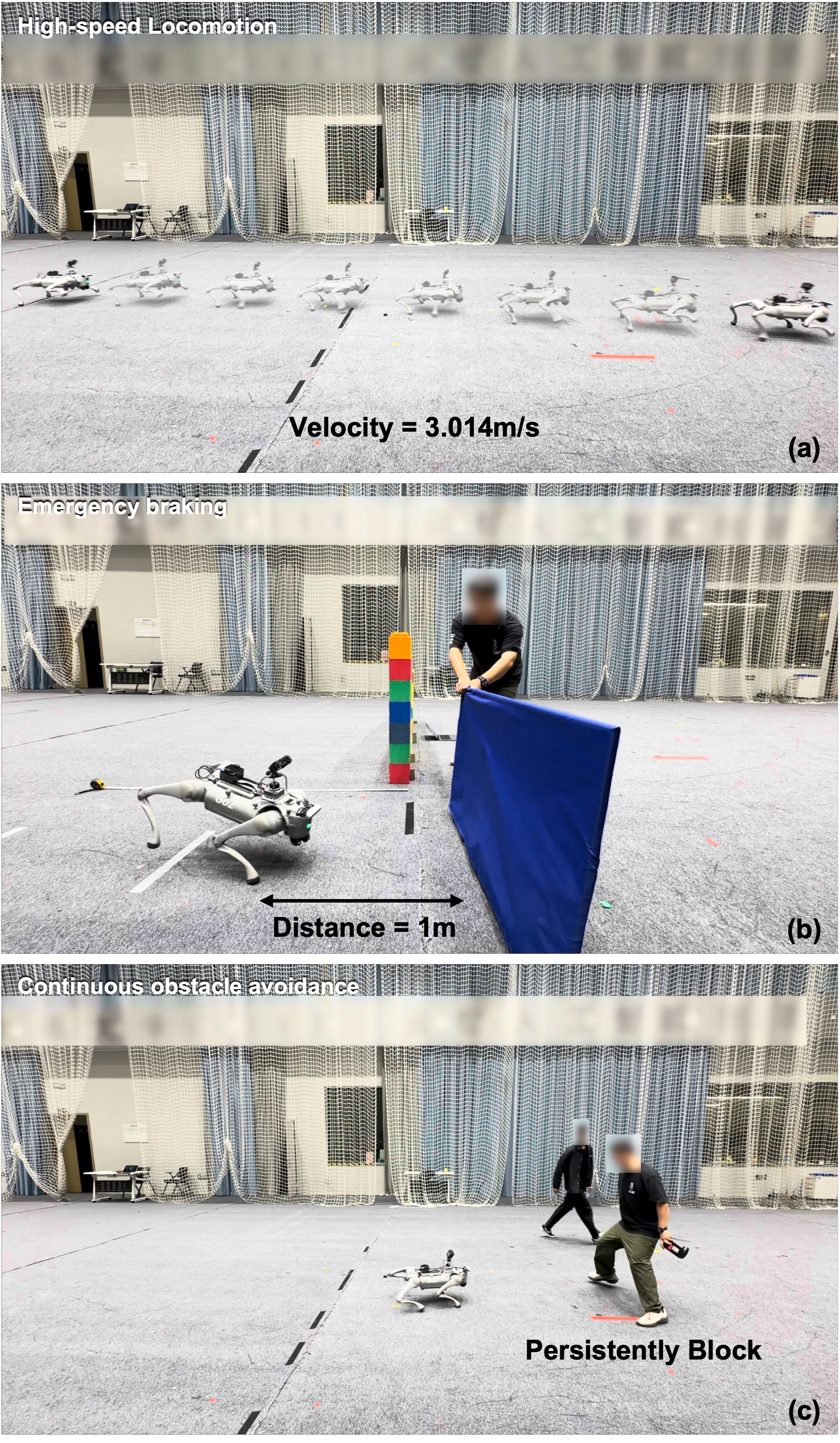}
    \caption{Snapshots of extreme performance and robustness experiments. (a) High-speed straight-line locomotion test, where the robot achieves a peak velocity of $3.014~\text{m/s}$ in an open environment. (b) Emergency braking test, illustrating the robot successfully stopping when an obstacle is suddenly thrown at a distance of $1~\text{m}$. (c) Continuous obstacle avoidance test, demonstrating the system's robustness against persistent dynamic interference from two operators.}
    \label{fig:extreme_tests}
\end{figure}

To evaluate the peak capabilities and robustness of VOP-Nav under limit conditions, we conducted a series of qualitative stress tests, as illustrated in Fig.~\ref{fig:extreme_tests}. The specific assessments are detailed as follows:

\begin{itemize}
    \item \textit{High-speed locomotion}: Conducted in an obstacle-free environment, this test aimed to determine the robot's maximum velocity. The experimental results recorded a peak instantaneous speed of $3.014~\text{m/s}$, verifying the controller's stability at high velocities.
    
    \item \textit{Emergency braking}: To assess rapid response capabilities, an operator hidden from view suddenly threw an obstacle into the robot's path. The results indicate that the robot successfully executed an emergency stop, even when the obstacle appeared at a critical distance of just $1~\text{m}$. Furthermore, the system autonomously resumed motion immediately after the obstacle was removed.
    
    \item \textit{Continuous obstacle avoidance}: In this scenario, two operators persistently blocked the robot's path to simulate aggressive interference. The system maintained stable operation throughout the disturbance without any collisions or falls, demonstrating high resilience against sustained dynamic perturbations.
\end{itemize}

In summary, these experiments demonstrate that VOP-Nav is capable of high-speed locomotion in open environments while exhibiting robust adaptability when facing extreme dynamic scenarios.

\section{Conclusion}
In this work, we presented VOP-Nav, a novel end-to-end navigation framework for quadruped robots that effectively balances safety and agility in dynamic and crowded environments. By combining Velocity Obstacle (VO) principles with reinforcement learning, our system overcomes the limitations of traditional methods that rely on fragile obstacle state estimation, as well as purely learning-based approaches that lack explicit safety guidance.

Central to our approach is VOP-Net, a perception module that implicitly encodes dynamic environmental constraints from multi-frame LiDAR data. Instead of reconstructing geometric shapes, VOP-Net directly regresses safe velocity intervals, providing a robust abstraction of navigable space. We show that using these predictions both as policy inputs during inference and as shaping terms in the reward function enables the navigation policy to learn effective and context-aware avoidance behaviors. Extensive evaluations in simulation demonstrate that VOP-Nav significantly outperforms state-of-the-art baselines in terms of success rates and collision avoidance, particularly in densely populated environments. Furthermore, zero-shot deployment on a Unitree Go2 quadruped robot confirms the robustness and efficiency of the system in real-world settings using only onboard sensors. 

Despite these results, the current system has two main limitations. First, VOP-Net operates on a 2D planar representation of the environment. While computationally efficient, this representation simplifies the complex 3D traversability unique to legged robots, potentially limiting performance in scenarios with overhanging structures or negative obstacles (e.g., ditches) that require volumetric perception. Second, robust operation in outdoor settings relies on the stability of {localization}. {In feature-poor environments or under extreme dynamic occlusion, LiDAR-inertial odometry may drift and introduce errors into the robot's linear-velocity and relative-goal-position inputs, thereby affecting the policy's navigation decisions.}

{VOP-Nav is particularly suitable for tasks where both agile motion and safe obstacle avoidance are important. Examples include time-critical tasks such as medical emergency response and disaster relief, as well as industrial tasks involving moving robots and human workers. VOP-Nav can operate independently and can also be naturally integrated into a full navigation stack as a local navigation module. Within such a stack, a higher-level planner provides VOP-Nav with sub-waypoints in the robot's local frame, each reachable along a straight path. This integration offers a practical way to extend the system toward longer-distance navigation in dense dynamic environments.}

Our findings suggest that encoding geometric priors into deep learning frameworks represents a promising direction. Future work will explore extending this velocity obstacle perception methodology to 3D volumetric spaces.

\bibliographystyle{IEEEtran}
\bibliography{voprl_refer}

@incollection{alonso-mora2013,
  title = {Optimal {{Reciprocal Collision Avoidance}} for {{Multiple Non-Holonomic Robots}}},
  booktitle = {Distributed {{Autonomous Robotic Systems}}},
  author = {{Alonso-Mora}, Javier and Breitenmoser, Andreas and Rufli, Martin and Beardsley, Paul and Siegwart, Roland},
  editor = {Martinoli, Alcherio and Mondada, Francesco and Correll, Nikolaus and Mermoud, Gr{\'e}gory and Egerstedt, Magnus and Hsieh, M. Ani and Parker, Lynne E. and St{\o}y, Kasper},
  year = {2013},
  volume = {83},
  pages = {203--216},
  publisher = {Springer Berlin Heidelberg},
  address = {Berlin, Heidelberg},
  doi = {10.1007/978-3-642-32723-0_15},
  urldate = {2025-04-29},
  copyright = {http://www.springer.com/tdm},
  isbn = {978-3-642-32722-3 978-3-642-32723-0},
  langid = {english}
}

@inproceedings{binpeng2020,
  title = {Learning {{Agile Robotic Locomotion Skills}} by {{Imitating Animals}}},
  booktitle = {Robotics: {{Science}} and {{Systems XVI}}},
  author = {Bin Peng, Xue and Coumans, Erwin and Zhang, Tingnan and Lee, Tsang-Wei and Tan, Jie and Levine, Sergey},
  year = {2020},
  month = jul,
  publisher = {{Robotics: Science and Systems Foundation}},
  doi = {10.15607/RSS.2020.XVI.064},
  urldate = {2025-05-08},
  isbn = {978-0-9923747-6-1},
  langid = {english}
}

@article{bledt2018,
  title = {{{MIT Cheetah}} 3: {{Design}} and {{Control}} of a {{Robust}}, {{Dynamic Quadruped Robot}}},
  shorttitle = {{{MIT Cheetah}} 3},
  author = {Bledt, Gerardo and Powell, Matthew J. and Katz, Benjamin and Di Carlo, Jared and Wensing, Patrick M. and Kim, Sangbae},
  year = {2018},
  month = oct,
  journal = {2018 IEEE/RSJ International Conference on Intelligent Robots and Systems (IROS)},
  pages = {2245--2252},
  publisher = {IEEE},
  address = {Madrid},
  doi = {10.1109/IROS.2018.8593885},
  urldate = {2025-05-08},
  isbn = {9781538680940}
}

@article{chen2024,
  title = {Reciprocal {{Velocity Obstacle Spatial-Temporal Network}} for {{Distributed Multirobot Navigation}}},
  author = {Chen, Lin and Wang, Yaonan and Miao, Zhiqiang and Feng, Mingtao and Zhou, Zhen and Wang, Hesheng and Wang, Danwei},
  year = {2024},
  month = nov,
  journal = {IEEE Transactions on Industrial Electronics},
  volume = {71},
  number = {11},
  pages = {14470--14480},
  issn = {1557-9948},
  doi = {10.1109/TIE.2024.3379630},
  urldate = {2025-04-29}
}

@inproceedings{cheng2024,
  title = {Extreme Parkour with Legged Robots},
  booktitle = {2024 {{IEEE International Conference}} on {{Robotics}} and {{Automation}} ({{ICRA}})},
  author = {Cheng, Xuxin and Shi, Kexin and Agarwal, Ananye and Pathak, Deepak},
  year = {2024},
  pages = {11443--11450},
  publisher = {IEEE},
  urldate = {2025-05-08}
}

@inproceedings{dicarlo2018,
  title = {Dynamic Locomotion in the Mit Cheetah 3 through Convex Model-Predictive Control},
  booktitle = {2018 {{IEEE}}/{{RSJ}} International Conference on Intelligent Robots and Systems ({{IROS}})},
  author = {Di Carlo, Jared and Wensing, Patrick M. and Katz, Benjamin and Bledt, Gerardo and Kim, Sangbae},
  year = {2018},
  pages = {1--9},
  publisher = {IEEE},
  urldate = {2025-05-08}
}

@article{fiorini1998,
  title = {Motion {{Planning}} in {{Dynamic Environments Using Velocity Obstacles}}},
  author = {Fiorini, Paolo and Shiller, Zvi},
  year = {1998},
  month = jul,
  journal = {The International Journal of Robotics Research},
  volume = {17},
  number = {7},
  pages = {760--772},
  issn = {0278-3649, 1741-3176},
  doi = {10.1177/027836499801700706},
  urldate = {2025-04-28},
  copyright = {https://journals.sagepub.com/page/policies/text-and-data-mining-license},
  langid = {english}
}

@inproceedings{fu2023,
  title = {Deep Whole-Body Control: Learning a Unified Policy for Manipulation and Locomotion},
  shorttitle = {Deep Whole-Body Control},
  booktitle = {Conference on {{Robot Learning}}},
  author = {Fu, Zipeng and Cheng, Xuxin and Pathak, Deepak},
  year = {2023},
  pages = {138--149},
  publisher = {PMLR},
  urldate = {2025-05-08}
}

@inproceedings{gopalakrishnan2017,
  title = {{{PRVO}}: {{Probabilistic Reciprocal Velocity Obstacle}} for Multi Robot Navigation under Uncertainty},
  shorttitle = {{{PRVO}}},
  booktitle = {2017 {{IEEE}}/{{RSJ International Conference}} on {{Intelligent Robots}} and {{Systems}} ({{IROS}})},
  author = {Gopalakrishnan, Bharath and Singh, Arun Kumar and Kaushik, Meha and Krishna, K. Madhava and Manocha, Dinesh},
  year = {2017},
  month = sep,
  pages = {1089--1096},
  issn = {2153-0866},
  doi = {10.1109/IROS.2017.8202279},
  urldate = {2025-04-28}
}

@article{grandia2023,
  title = {Perceptive Locomotion through Nonlinear Model-Predictive Control},
  author = {Grandia, Ruben and Jenelten, Fabian and Yang, Shaohui and Farshidian, Farbod and Hutter, Marco},
  year = {2023},
  journal = {IEEE Transactions on Robotics},
  volume = {39},
  number = {5},
  pages = {3402--3421},
  publisher = {IEEE},
  urldate = {2025-05-08}
}

@article{haarnoja2024,
  title = {Learning Agile Soccer Skills for a Bipedal Robot with Deep Reinforcement Learning},
  author = {Haarnoja, Tuomas and Moran, Ben and Lever, Guy and Huang, Sandy H. and Tirumala, Dhruva and Humplik, Jan and Wulfmeier, Markus and Tunyasuvunakool, Saran and Siegel, Noah Y. and Hafner, Roland and Bloesch, Michael and Hartikainen, Kristian and Byravan, Arunkumar and Hasenclever, Leonard and Tassa, Yuval and Sadeghi, Fereshteh and Batchelor, Nathan and Casarini, Federico and Saliceti, Stefano and Game, Charles and Sreendra, Neil and Patel, Kushal and Gwira, Marlon and Huber, Andrea and Hurley, Nicole and Nori, Francesco and Hadsell, Raia and Heess, Nicolas},
  year = {2024},
  month = apr,
  journal = {Science Robotics},
  volume = {9},
  number = {89},
  pages = {eadi8022},
  issn = {2470-9476},
  doi = {10.1126/scirobotics.adi8022},
  urldate = {2025-05-08},
  langid = {english}
}

@article{han2022,
  title = {Reinforcement {{Learned Distributed Multi-Robot Navigation With Reciprocal Velocity Obstacle Shaped Rewards}}},
  author = {Han, Ruihua and Chen, Shengduo and Wang, Shuaijun and Zhang, Zeqing and Gao, Rui and Hao, Qi and Pan, Jia},
  year = {2022},
  month = jul,
  journal = {IEEE Robotics and Automation Letters},
  volume = {7},
  number = {3},
  pages = {5896--5903},
  issn = {2377-3766},
  doi = {10.1109/LRA.2022.3161699},
  urldate = {2024-04-24}
}

@inproceedings{he2024,
  title = {Agile {{But Safe}}: {{Learning Collision-Free High-Speed Legged Locomotion}}},
  shorttitle = {Agile {{But Safe}}},
  booktitle = {Robotics: {{Science}} and {{Systems XX}}},
  author = {He, Tairan and Zhang, Chong and Xiao, Wenli and He, Guanqi and Liu, Changliu and Shi, Guanya},
  year = 2024,
  month = jul,
  publisher = {{Robotics: Science and Systems Foundation}},
  doi = {10.15607/RSS.2024.XX.059},
  urldate = {2026-01-13},
  isbn = {979-8-9902848-0-7},
  langid = {english}
}

@article{huajian2024,
  title = {Sample-{{Efficient Learning-Based Dynamic Environment Navigation With Transferring Experience From Optimization-Based Planner}}},
  author = {Huajian, Liu and Wei, Dong and Shouren, Mao and Chao, Wang and Yongzhuo, Gao},
  year = {2024},
  month = aug,
  journal = {IEEE Robotics and Automation Letters},
  volume = {9},
  number = {8},
  pages = {7055--7062},
  issn = {2377-3766},
  doi = {10.1109/LRA.2024.3412610},
  urldate = {2025-04-29}
}

@article{hwangbo2019,
  title = {Learning Agile and Dynamic Motor Skills for Legged Robots},
  author = {Hwangbo, Jemin and Lee, Joonho and Dosovitskiy, Alexey and Bellicoso, Dario and Tsounis, Vassilios and Koltun, Vladlen and Hutter, Marco},
  year = {2019},
  month = jan,
  journal = {Science Robotics},
  volume = {4},
  number = {26},
  pages = {eaau5872},
  issn = {2470-9476},
  doi = {10.1126/scirobotics.aau5872},
  urldate = {2025-05-08},
  langid = {english}
}

@article{ji2022,
  title = {Concurrent Training of a Control Policy and a State Estimator for Dynamic and Robust Legged Locomotion},
  author = {Ji, Gwanghyeon and Mun, Juhyeok and Kim, Hyeongjun and Hwangbo, Jemin},
  year = {2022},
  journal = {IEEE Robotics and Automation Letters},
  volume = {7},
  number = {2},
  pages = {4630--4637},
  publisher = {IEEE},
  urldate = {2025-05-08}
}

@misc{kim2019,
  title = {Highly {{Dynamic Quadruped Locomotion}} via {{Whole-Body Impulse Control}} and {{Model Predictive Control}}},
  author = {Kim, Donghyun and Carlo, Jared Di and Katz, Benjamin and Bledt, Gerardo and Kim, Sangbae},
  year = {2019},
  month = sep,
  number = {arXiv:1909.06586},
  primaryclass = {cs},
  publisher = {arXiv},
  doi = {10.48550/arXiv.1909.06586},
  urldate = {2025-05-08},
  archiveprefix = {arXiv}
}

@misc{lee2019,
  title = {Robust {{Recovery Controller}} for a {{Quadrupedal Robot}} Using {{Deep Reinforcement Learning}}},
  author = {Lee, Joonho and Hwangbo, Jemin and Hutter, Marco},
  year = {2019},
  month = jan,
  number = {arXiv:1901.07517},
  primaryclass = {cs},
  publisher = {arXiv},
  doi = {10.48550/arXiv.1901.07517},
  urldate = {2025-05-08},
  archiveprefix = {arXiv}
}

@inproceedings{liu2024,
  title = {Visual {{Whole-Body Control}} for {{Legged Loco-Manipulation}}},
  booktitle = {8th {{Annual Conference}} on {{Robot Learning}}},
  author = {Liu, Minghuan and Chen, Zixuan and Cheng, Xuxin and Ji, Yandong and Qiu, Ri-Zhao and Yang, Ruihan and Wang, Xiaolong},
  year = 2024,
  month = sep,
  urldate = {2026-01-13},
  langid = {english}
}

@misc{liu2025,
  title = {Discrete-{{Time Hybrid Automata Learning}}: {{Legged Locomotion Meets Skateboarding}}},
  shorttitle = {Discrete-{{Time Hybrid Automata Learning}}},
  author = {Liu, Hang and Teng, Sangli and Liu, Ben and Zhang, Wei and Ghaffari, Maani},
  year = {2025},
  month = apr,
  number = {arXiv:2503.01842},
  primaryclass = {cs},
  publisher = {arXiv},
  doi = {10.48550/arXiv.2503.01842},
  urldate = {2025-05-08},
  archiveprefix = {arXiv}
}

@inproceedings{ma2023,
  title = {Learning Arm-Assisted Fall Damage Reduction and Recovery for Legged Mobile Manipulators},
  booktitle = {2023 {{IEEE International Conference}} on {{Robotics}} and {{Automation}} ({{ICRA}})},
  author = {Ma, Yuntao and Farshidian, Farbod and Hutter, Marco},
  year = {2023},
  pages = {12149--12155},
  publisher = {IEEE},
  urldate = {2025-05-08}
}

@inproceedings{macenski2020,
  title = {The Marathon 2: {{A}} Navigation System},
  shorttitle = {The Marathon 2},
  booktitle = {2020 {{IEEE}}/{{RSJ International Conference}} on {{Intelligent Robots}} and {{Systems}} ({{IROS}})},
  author = {Macenski, Steve and Mart{\'i}n, Francisco and White, Ruffin and Clavero, Jonatan Gin{\'e}s},
  year = {2020},
  pages = {2718--2725},
  publisher = {IEEE},
  urldate = {2025-05-08}
}

@article{makoviychuk2021,
  title = {Isaac {{Gym}}: {{High Performance GPU-Based Physics Simulation For Robot Learning}}},
  shorttitle = {Isaac {{Gym}}},
  author = {Makoviychuk, Viktor and Wawrzyniak, Lukasz and Guo, Yunrong and Lu, Michelle and Storey, Kier and Macklin, M. and Hoeller, David and Rudin, N. and Allshire, Arthur and Handa, Ankur and State, Gavriel},
  year = {2021},
  month = aug,
  journal = {ArXiv},
  urldate = {2025-05-08}
}

@inproceedings{marder-eppstein2010,
  title = {The Office Marathon: {{Robust}} Navigation in an Indoor Office Environment},
  shorttitle = {The Office Marathon},
  booktitle = {2010 {{IEEE}} International Conference on Robotics and Automation},
  author = {{Marder-Eppstein}, Eitan and Berger, Eric and Foote, Tully and Gerkey, Brian and Konolige, Kurt},
  year = {2010},
  pages = {300--307},
  publisher = {IEEE},
  urldate = {2025-05-08}
}

@article{margolis2024,
  title = {Rapid Locomotion via Reinforcement Learning},
  author = {Margolis, Gabriel B. and Yang, Ge and Paigwar, Kartik and Chen, Tao and Agrawal, Pulkit},
  year = {2024},
  month = apr,
  journal = {The International Journal of Robotics Research},
  volume = {43},
  number = {4},
  pages = {572--587},
  issn = {0278-3649, 1741-3176},
  doi = {10.1177/02783649231224053},
  urldate = {2025-05-08},
  langid = {english}
}

@article{neunert2018,
  title = {Whole-Body Nonlinear Model Predictive Control through Contacts for Quadrupeds},
  author = {Neunert, Michael and St{\"a}uble, Markus and Giftthaler, Markus and Bellicoso, Carmine D. and Carius, Jan and Gehring, Christian and Hutter, Marco and Buchli, Jonas},
  year = {2018},
  journal = {IEEE Robotics and Automation Letters},
  volume = {3},
  number = {3},
  pages = {1458--1465},
  publisher = {IEEE},
  urldate = {2025-05-08}
}

@article{qin2024,
  title = {{{SRL-ORCA}}: {{A Socially Aware Multi-Agent Mapless Navigation Algorithm}} in {{Complex Dynamic Scenes}}},
  shorttitle = {{{SRL-ORCA}}},
  author = {Qin, Jianmin and Qin, Jiahu and Qiu, Jiaxin and Liu, Qingchen and Li, Man and Ma, Qichao},
  year = {2024},
  month = jan,
  journal = {IEEE Robotics and Automation Letters},
  volume = {9},
  number = {1},
  pages = {143--150},
  issn = {2377-3766},
  doi = {10.1109/LRA.2023.3331621},
  urldate = {2025-04-29}
}

@misc{schulman2017,
  title = {Proximal {{Policy Optimization Algorithms}}},
  author = {Schulman, John and Wolski, Filip and Dhariwal, Prafulla and Radford, Alec and Klimov, Oleg},
  year = {2017},
  month = aug,
  number = {arXiv:1707.06347},
  primaryclass = {cs},
  publisher = {arXiv},
  doi = {10.48550/arXiv.1707.06347},
  urldate = {2025-05-12},
  archiveprefix = {arXiv}
}

@article{snape2011,
  title = {The {{Hybrid Reciprocal Velocity Obstacle}}},
  author = {Snape, Jamie and van den Berg, Jur and Guy, Stephen J. and Manocha, Dinesh},
  year = {2011},
  month = aug,
  journal = {IEEE Transactions on Robotics},
  volume = {27},
  number = {4},
  pages = {696--706},
  issn = {1941-0468},
  doi = {10.1109/TRO.2011.2120810},
  urldate = {2025-04-28}
}

@inproceedings{todorov2012,
  title = {Mujoco: {{A}} Physics Engine for Model-Based Control},
  shorttitle = {Mujoco},
  booktitle = {2012 {{IEEE}}/{{RSJ}} International Conference on Intelligent Robots and Systems},
  author = {Todorov, Emanuel and Erez, Tom and Tassa, Yuval},
  year = {2012},
  pages = {5026--5033},
  publisher = {IEEE},
  urldate = {2025-05-08}
}

@inproceedings{vandenberg2008,
  title = {Reciprocal {{Velocity Obstacles}} for Real-Time Multi-Agent Navigation},
  booktitle = {2008 {{IEEE International Conference}} on {{Robotics}} and {{Automation}}},
  author = {{van den Berg}, Jur and Lin, Ming and Manocha, Dinesh},
  year = {2008},
  month = may,
  pages = {1928--1935},
  issn = {1050-4729},
  doi = {10.1109/ROBOT.2008.4543489},
  urldate = {2025-04-28}
}

@inproceedings{vandenberg2011,
  title = {Reciprocal N-{{Body Collision Avoidance}}},
  booktitle = {Robotics {{Research}}},
  author = {{van den Berg}, Jur and Guy, Stephen J. and Lin, Ming and Manocha, Dinesh},
  editor = {Pradalier, C{\'e}dric and Siegwart, Roland and Hirzinger, Gerhard},
  year = {2011},
  pages = {3--19},
  publisher = {Springer},
  address = {Berlin, Heidelberg},
  doi = {10.1007/978-3-642-19457-3_1},
  isbn = {978-3-642-19457-3},
  langid = {english}
}

@article{xie2023,
  title = {{{DRL-VO}}: {{Learning}} to {{Navigate Through Crowded Dynamic Scenes Using Velocity Obstacles}}},
  shorttitle = {{{DRL-VO}}},
  author = {Xie, Zhanteng and Dames, Philip},
  year = {2023},
  month = aug,
  journal = {IEEE Transactions on Robotics},
  volume = {39},
  number = {4},
  primaryclass = {cs},
  pages = {2700--2719},
  issn = {1552-3098, 1941-0468},
  doi = {10.1109/TRO.2023.3257549},
  urldate = {2024-08-28},
  archiveprefix = {arXiv}
}

@article{xu2025,
  title = {{{NavRL}}: {{Learning Safe Flight}} in {{Dynamic Environments}}},
  shorttitle = {{{NavRL}}},
  author = {Xu, Zhefan and Han, Xinming and Shen, Haoyu and Jin, Hanyu and Shimada, Kenji},
  year = {2025},
  month = apr,
  journal = {IEEE Robotics and Automation Letters},
  volume = {10},
  number = {4},
  pages = {3668--3675},
  issn = {2377-3766},
  doi = {10.1109/LRA.2025.3546069},
  urldate = {2025-04-30}
}

@article{zhang2024,
  title={Learning vision-based agile flight via differentiable physics},
  author={Zhang, Yuang and Hu, Yu and Song, Yunlong and Zou, Danping and Lin, Weiyao},
  journal={Nature Machine Intelligence},
  pages={1--13},
  year={2025},
  publisher={Nature Publishing Group UK London}
}

@article{zhou2019,
  title = {Robust and {{Efficient Quadrotor Trajectory Generation}} for {{Fast Autonomous Flight}}},
  author = {Zhou, Boyu and Gao, Fei and Wang, Luqi and Liu, Chuhao and Shen, Shaojie},
  year = {2019},
  month = oct,
  journal = {IEEE Robotics and Automation Letters},
  volume = {4},
  number = {4},
  pages = {3529--3536},
  issn = {2377-3766, 2377-3774},
  doi = {10.1109/LRA.2019.2927938},
  urldate = {2025-05-08},
  copyright = {https://ieeexplore.ieee.org/Xplorehelp/downloads/license-information/IEEE.html}
}

@article{zhou2022,
  title = {Navigating {{Robots}} in {{Dynamic Environment With Deep Reinforcement Learning}}},
  author = {Zhou, Zhiqian and Zeng, Zhiwen and Lang, Lin and Yao, Weijia and Lu, Huimin and Zheng, Zhiqiang and Zhou, Zongtan},
  year = {2022},
  month = dec,
  journal = {IEEE Transactions on Intelligent Transportation Systems},
  volume = {23},
  number = {12},
  pages = {25201--25211},
  issn = {1558-0016},
  doi = {10.1109/TITS.2022.3213604},
  urldate = {2025-04-29}
}

@article{zhou2023,
  title = {A Safe Reinforcement Learning Approach for Autonomous Navigation of Mobile Robots in Dynamic Environments},
  author = {Zhou, Zhiqian and Ren, Junkai and Zeng, Zhiwen and Xiao, Junhao and Zhang, Xinglong and Guo, Xian and Zhou, Zongtan and Lu, Huimin},
  year = {2023},
  month = oct,
  journal = {CAAI Transactions on Intelligence Technology},
  pages = {cit2.12269},
  issn = {2468-2322, 2468-2322},
  doi = {10.1049/cit2.12269},
  urldate = {2025-04-29},
  langid = {english}
}

@inproceedings{zhuang2023,
  title = {Robot {{Parkour Learning}}},
  booktitle = {7th {{Annual Conference}} on {{Robot Learning}}},
  author = {Zhuang, Ziwen and Fu, Zipeng and Wang, Jianren and Atkeson, Christopher G. and Schwertfeger, S{\"o}ren and Finn, Chelsea and Zhao, Hang},
  year = 2023,
  month = aug,
  urldate = {2026-01-13},
  langid = {english}
}

@inproceedings{zhuang2024,
  title = {Humanoid {{Parkour Learning}}},
  booktitle = {8th {{Annual Conference}} on {{Robot Learning}}},
  author = {Zhuang, Ziwen and Yao, Shenzhe and Zhao, Hang},
  year = 2024,
  month = sep,
  urldate = {2026-01-13},
  langid = {english}
}

@inproceedings{zhang2024a,
  title = {Learning {{Agile Locomotion}} on {{Risky Terrains}}},
  booktitle = {2024 {{IEEE}}/{{RSJ International Conference}} on {{Intelligent Robots}} and {{Systems}} ({{IROS}})},
  author = {Zhang, Chong and Rudin, Nikita and Hoeller, David and Hutter, Marco},
  year = 2024,
  month = oct,
  pages = {11864--11871},
  issn = {2153-0866},
  doi = {10.1109/IROS58592.2024.10801909},
  urldate = {2026-01-13}
}

@inproceedings{chen2019a,
  title = {Crowd-{{Robot Interaction}}: {{Crowd-Aware Robot Navigation With Attention-Based Deep Reinforcement Learning}}},
  shorttitle = {Crowd-{{Robot Interaction}}},
  booktitle = {2019 {{International Conference}} on {{Robotics}} and {{Automation}} ({{ICRA}})},
  author = {Chen, Changan and Liu, Yuejiang and Kreiss, Sven and Alahi, Alexandre},
  year = 2019,
  month = may,
  pages = {6015--6022},
  issn = {2577-087X},
  doi = {10.1109/ICRA.2019.8794134},
  urldate = {2026-01-13}
}

@misc{fan2018,
  title = {{{CrowdMove}}: {{Autonomous Mapless Navigation}} in {{Crowded Scenarios}}},
  shorttitle = {{{CrowdMove}}},
  author = {Fan, Tingxiang and Cheng, Xinjing and Pan, Jia and Manocha, Dinesh and Yang, Ruigang},
  year = {2018},
  month = jul,
  number = {arXiv:1807.07870},
  primaryclass = {cs},
  publisher = {arXiv},
  doi = {10.48550/arXiv.1807.07870},
  urldate = {2024-02-20},
  archiveprefix = {arXiv}
}

@article{liu2025a,
  title = {{{HEIGHT}}: {{Heterogeneous Interaction Graph Transformer}} for {{Robot Navigation}} in {{Crowded}} and {{Constrained Environments}}},
  shorttitle = {{{HEIGHT}}},
  author = {Liu, Shuijing and Xia, Haochen and Pouria, Fatemeh Cheraghi and Hong, Kaiwen and Chakraborty, Neeloy and Hu, Zichao and Biswas, Joydeep and {Driggs-Campbell}, Katherine},
  year = 2025,
  journal = {IEEE Transactions on Automation Science and Engineering},
  pages = {1--1},
  issn = {1558-3783},
  doi = {10.1109/TASE.2025.3646588},
  urldate = {2026-01-13}
}

@inproceedings{medina2023,
  title = {Human-{{Aware Navigation}} in {{Crowded Environments Using Adaptive Proxemic Area}} and {{Group Detection}}},
  booktitle = {2023 {{IEEE}}/{{RSJ International Conference}} on {{Intelligent Robots}} and {{Systems}} ({{IROS}})},
  author = {{Medina-S{\'a}nchez}, Carlos and Janzon, Simon and Zella, Matteo and Capit{\'a}n, Jes{\'u}s and Marr{\'o}n, Pedro J.},
  year = {2023},
  month = oct,
  pages = {6741--6748},
  publisher = {IEEE},
  address = {Detroit, MI, USA},
  doi = {10.1109/IROS55552.2023.10342385},
  urldate = {2024-01-05},
  isbn = {978-1-6654-9190-7},
  langid = {english}
}

@misc{zhou2023a,
  title = {Robot {{Crowd Navigation}} in {{Dynamic Environment}} with {{Offline Reinforcement Learning}}},
  author = {Zhou, Shuai and Fu, Hao and He, Haodong and Liu, Wei},
  year = {2023},
  month = dec,
  number = {arXiv:2312.11032},
  primaryclass = {cs},
  publisher = {arXiv},
  doi = {10.48550/arXiv.2312.11032},
  urldate = {2025-07-01},
  archiveprefix = {arXiv}
}

@misc{helbing1998,
  title = {Social {{Force Model}} for {{Pedestrian Dynamics}}},
  author = {Helbing, Dirk and Molnar, Peter},
  year = {1998},
  month = may,
  doi = {10.1103/PhysRevE.51.4282},
  urldate = {2025-05-29},
  archiveprefix = {arXiv}
}

@inproceedings{ahmed2022,
  title = {End-to-{{End Mobile Robot Navigation}} Using a {{Residual Deep Reinforcement Learning}} in {{Dynamic Human Environments}}},
  booktitle = {2022 18th {{IEEE}}/{{ASME International Conference}} on {{Mechatronic}} and {{Embedded Systems}} and {{Applications}} ({{MESA}})},
  author = {Ahmed, Abdullah and Mohammad, Yasser F. O. and Parque, Victor and {El-Hussieny}, Haitham and Ahmed, Sabah},
  year = {2022},
  month = nov,
  pages = {1--6},
  publisher = {IEEE},
  address = {Taipei, Taiwan},
  doi = {10.1109/MESA55290.2022.10004394},
  urldate = {2024-01-04},
  isbn = {978-1-6654-5570-1},
  langid = {english}
}

@inproceedings{pfeiffer2017,
  title = {From {{Perception}} to {{Decision}}: {{A Data-driven Approach}} to {{End-to-end Motion Planning}} for {{Autonomous Ground Robots}}},
  shorttitle = {From {{Perception}} to {{Decision}}},
  booktitle = {2017 {{IEEE International Conference}} on {{Robotics}} and {{Automation}} ({{ICRA}})},
  author = {Pfeiffer, Mark and Schaeuble, Michael and Nieto, Juan and Siegwart, Roland and Cadena, Cesar},
  year = {2017},
  month = may,
  primaryclass = {cs},
  pages = {1527--1533},
  doi = {10.1109/ICRA.2017.7989182},
  urldate = {2025-06-22},
  archiveprefix = {arXiv}
}

@article{xu2022,
  title = {{{FAST-LIO2}}: {{Fast Direct LiDAR-Inertial Odometry}}},
  shorttitle = {{{FAST-LIO2}}},
  author = {Xu, Wei and Cai, Yixi and He, Dongjiao and Lin, Jiarong and Zhang, Fu},
  year = 2022,
  month = aug,
  journal = {IEEE Transactions on Robotics},
  volume = {38},
  number = {4},
  pages = {2053--2073},
  issn = {1941-0468},
  doi = {10.1109/TRO.2022.3141876},
  urldate = {2026-07-14}
}

@article{fox1997,
  title = {The Dynamic Window Approach to Collision Avoidance},
  author = {Fox, D. and Burgard, W. and Thrun, S.},
  year = {1997},
  month = mar,
  journal = {IEEE Robotics \& Automation Magazine},
  volume = {4},
  number = {1},
  pages = {23--33},
  issn = {1558-223X},
  doi = {10.1109/100.580977},
  urldate = {2025-06-29}
}

@article{martinez-baselga2025,
  title = {{{AVOCADO}}: {{Adaptive Optimal Collision Avoidance Driven}} by {{Opinion}}},
  shorttitle = {{{AVOCADO}}},
  author = {{Martinez-Baselga}, Diego and Sebasti{\'a}n, Eduardo and Montijano, Eduardo and Riazuelo, Luis and Sag{\"u}{\'e}s, Carlos and Montano, Luis},
  year = 2025,
  journal = {IEEE Transactions on Robotics},
  volume = {41},
  pages = {2495--2511},
  issn = {1941-0468},
  doi = {10.1109/TRO.2025.3552350},
  urldate = {2026-01-08}
}

@inproceedings{xiao2025,
  title = {{{DGVO}}: {{A Dynamically Constrained Gradient Velocity Obstacle Approach}} for {{Mobile Robots}} in {{Dynamic Environments}}},
  shorttitle = {{{DGVO}}},
  booktitle = {2025 {{IEEE}}/{{RSJ International Conference}} on {{Intelligent Robots}} and {{Systems}} ({{IROS}})},
  author = {Xiao, Bowen and Zhang, Bo and Zhang, Danyu and Xie, Peiyan and Wang, Xinyu and Li, Ruocheng},
  year = 2025,
  month = oct,
  pages = {19675--19682},
  issn = {2153-0866},
  doi = {10.1109/IROS60139.2025.11247136},
  urldate = {2026-01-08}
}

@article{yuan2025reasan,
  title = {{{REASAN}}: {{Learning Reactive Safe Navigation}} for {{Legged Robots}}},
  author = {Yuan, Qihao and Cao, Ziyu and Cao, Ming and Li, Kailai},
  year = {2025},
  journal = {arXiv preprint arXiv:2512.09537}
}

\end{document}